\documentclass[final,5p,times,twocolumn]{elsarticle}

\usepackage{amsmath,amsfonts,amssymb}
\usepackage{graphicx}
\usepackage{subcaption}
\usepackage{booktabs}
\usepackage{bm}
\usepackage{multirow}
\usepackage{threeparttable}
\usepackage{hyperref}
\usepackage{color}
\usepackage[justification=raggedright,singlelinecheck=false]{caption}

\hypersetup{
    colorlinks=true,
    linkcolor=Maroon,
    citecolor=Maroon,
    urlcolor=Maroon
}

\journal{ }

\begin{document}

\begin{frontmatter}

% -----------------------------------------------------
% TITLE
% -----------------------------------------------------
%\title{Bayesian Optimization over Kernel-of-Kernels via Gaussian Processes on MDS-Embedded Kernel Manifolds}

\title{The Kernel Manifold: A Geometric Approach to Gaussian Process Model Selection}

\author[1]{Md Shafiqul Islam\corref{cor1}}
\ead{shafiq@tamu.edu}
\cortext[cor1]{Corresponding author}

\author[1]{Shakti Prasad Padhy}
\ead{shaktippadhy@tamu.edu}

\author[2]{Douglas Allaire}
\ead{dallaire@tamu.edu}

\author[1,2]{Raymundo Arr\'oyave}
\ead{raymundo.arroyave@tamu.edu}

\affiliation[1]{organization={Department of Materials Science and Engineering, Texas A\&M University},
                city={College Station},
                postcode={77843}, 
                state={TX},
                country={USA}}

\affiliation[2]{organization={J. Mike Walker '66 Department of Mechanical Engineering, Texas A\&M University},
                city={College Station},
                postcode={77843}, 
                state={TX},
                country={USA}}

% -----------------------------------------------------
% ABSTRACT
% -----------------------------------------------------
\begin{abstract}
Gaussian Process (GP) regression is a powerful nonparametric Bayesian framework, but its performance depends critically on the choice of covariance kernel. Selecting an appropriate kernel is therefore central to model quality, yet remains one of the most challenging and computationally expensive steps in probabilistic modeling. We present a Bayesian optimization framework built on kernel-of-kernels geometry, using expected divergence-based distances between GP priors to explore kernel space efficiently. A multidimensional scaling (MDS) embedding of this distance matrix maps a discrete kernel library into a continuous Euclidean manifold, enabling smooth BO. In this formulation, the input space comprises kernel compositions, the objective is the log marginal likelihood, and featurization is given by the MDS coordinates. When the divergence yields a valid metric, the embedding preserves geometry and produces a stable BO landscape. We demonstrate the approach on synthetic benchmarks, real-world time-series datasets, and an additive manufacturing case study predicting melt-pool geometry, achieving superior predictive accuracy and uncertainty calibration relative to baselines including Large Language Model (LLM)-guided search. This framework establishes a reusable probabilistic geometry for kernel search, with direct relevance to GP modeling and deep kernel learning.
\end{abstract}


\end{frontmatter}

% -----------------------------------------------------
% INTRODUCTION
% -----------------------------------------------------
\section{Introduction}

Gaussian process (GP) regression provides a principled framework for probabilistic learning. By treating unknown functions as draws from a GP prior, it yields closed-form posterior inference, calibrated uncertainty, and data-driven hyperparameter learning. The covariance kernel encodes prior assumptions about smoothness, periodicity, and compositional structure. For example, the squared-exponential (SE) kernel enforces infinitely differentiable smoothness, Mat\'ern kernels allow rougher behavior, and periodic or rational quadratic (RQ) kernels capture recurring patterns and multi-scale structure.

However, kernel selection remains a critical bottleneck. Standard approaches---hand-designing kernels or searching small libraries of base kernels and their sums/products---are restrictive. Complex data often require compositional kernels formed by algebraic combinations (e.g., $k = k_{\mathrm{SE}} + k_{\mathrm{PER}} \times k_{\mathrm{RQ}}$), which rapidly leads to a combinatorial explosion that makes exhaustive search impractical. Moreover, kernel hyperparameters (length scales, periodicities, amplitudes) can dramatically alter the induced GP prior, so small perturbations can make similar symbolic kernels yield very different predictive distributions, limiting purely symbolic search.

In this work, we cast \emph{kernel discovery as a geometric problem} over Gaussian process priors. Rather than manipulating symbolic expressions or individual hyperparameters, we define an \emph{expected-divergence distance} between GP priors by integrating probabilistic divergences over kernel hyperparameters. This induces a task-aware \emph{kernel-of-kernels geometry} that compares kernels by the stochastic function distributions they generate, rather than by algebraic form.

\begin{figure}[h]
  \centering
  \includegraphics[width=0.99\columnwidth]{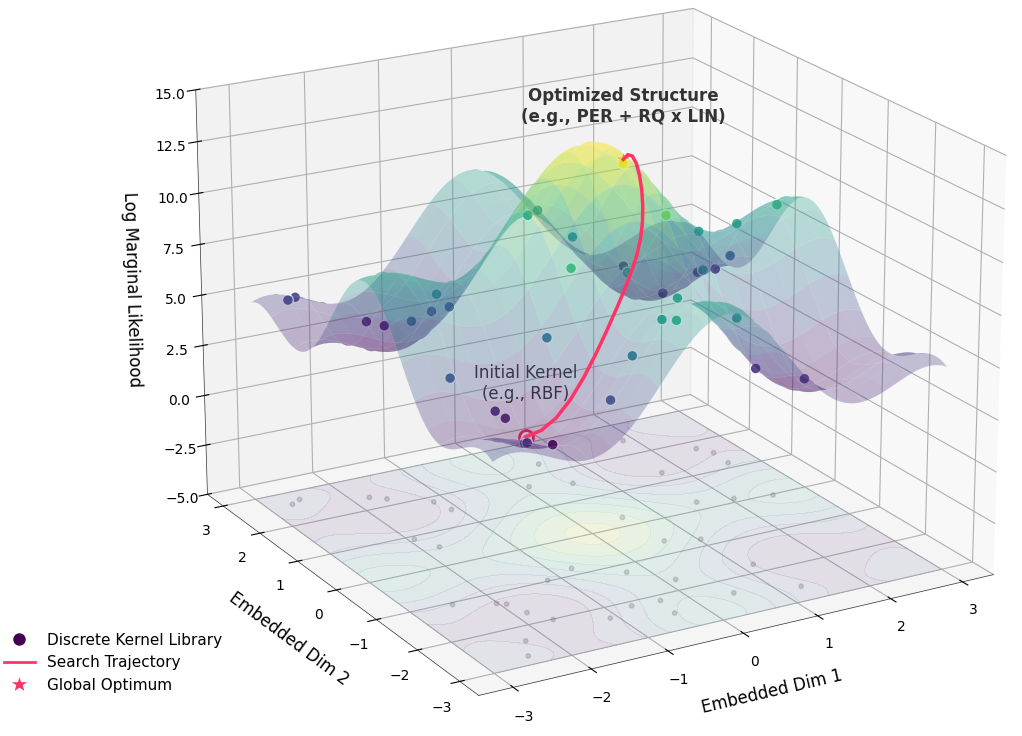}
  \caption{Conceptual overview of BO for kernel selection on the kernel‑of‑kernels manifold. Discrete kernel
  candidates are embedded into a continuous space, and BO traverses the log‑marginal‑likelihood
  landscape from an initial kernel toward an optimized structure.}
  \label{fig:schematic}
\end{figure}

To make this geometry operational for optimization, we embed the resulting kernel--kernel distance matrix into a continuous Euclidean space using \emph{multidimensional scaling} (MDS). This embedding assigns fixed coordinates to a discrete library of compositional kernels, transforming kernel selection into a continuous optimization problem. Bayesian optimization is then performed directly on the embedded manifold, while evaluation remains restricted to the original discrete kernel candidates. In this formulation, the embedded kernel coordinates define the input space, the objective is the log marginal likelihood, and similarity in kernel geometry is captured implicitly by the surrogate covariance function. A conceptual overview is shown in Fig.~\ref{fig:schematic}.

Early attempts to broaden kernel choice for Gaussian processes framed kernel discovery as search over a compositional grammar closed under sums and products of base kernels. Duvenaud et al.~\cite{duvenaud2013structure} proposed a greedy procedure that scores candidate structures with an approximate marginal likelihood, showing that learned composite kernels yield interpretable decompositions and strong extrapolation on time-series data. Their method iteratively expands grammar expressions and uses the Bayesian Information Criterion (BIC) as a tractable, penalized-likelihood proxy during search.

Building on the grammar-based perspective, Malkomes, Schaff, and Garnett~\cite{malkomes2016bayesian} reframed kernel selection as Bayesian optimization over model space. Rather than greedy enumeration, their method optimizes the marginal likelihood over a large (in principle unbounded) kernel space using a kernel between models that reflects how well they explain the dataset. This kernel-of-models view reduces the number of expensive marginal-likelihood evaluations while remaining data-adaptive.

A different line of work uses likelihood-free Bayesian inference for kernel selection. Abdessalem et al.~\cite{abdessalem2017automatic} employed Approximate Bayesian Computation with Sequential Monte Carlo (ABC-SMC) to jointly select kernel structures and infer hyperparameters by comparing simulated and observed data via a user-defined distance metric. By bypassing explicit likelihoods, the approach can transition between kernels of different dimensionality and remains robust when Gaussian assumptions are questionable or evidence is hard to evaluate.

Recent work explores large-language-model (LLM)--assisted adaptive kernel evolution~\cite{suwandi2025adaptive}. The CAKE framework uses an LLM to propose kernels from a compositional grammar and ranks candidates with a BIC--acquisition hybrid (BAKER) to balance fit and utility at each BO iteration. Reported results show improvements over fixed-kernel and adaptive baselines across hyperparameter optimization, control, and photonics tasks, suggesting that LLM priors can accelerate kernel search in few-shot regimes.

The approaches above address kernel discovery by operating directly on symbolic kernel structures or hyperparameterized families. They either search compositional spaces with likelihood-based scores (greedy or BO), perform likelihood-free inference over kernels and hyperparameters (ABC-SMC), or enlist LLMs to generate symbolic candidates during optimization. Despite differences, they treat kernels as discrete objects whose structure or parameters are manipulated during search.

A central challenge is defining meaningful similarity measures between kernels. Probabilistic distances compare kernels at the level of induced function distributions rather than symbolic form. The Hellinger distance is symmetric and bounded and has been used in model-space BO settings \cite{malkomes2016bayesian}. In practice, however, Hellinger distances between GP priors often concentrate near one, yielding low-contrast distance matrices that compress the geometry and complicate embedding and optimization.

Addressing these issues requires examining the divergence used to construct kernel geometry and transforming the resulting distance matrix to recover variation and curvature. We then develop a principled framework for kernel discovery from this geometric perspective: we analyze probabilistic divergences between GP priors, study the geometric and spectral properties they induce, and identify transformations that enable stable Euclidean embeddings via multidimensional scaling. Building on this representation, we perform Bayesian optimization on the embedded kernel manifold while restricting evaluation to discrete kernel candidates, and we evaluate the approach on synthetic benchmarks and real-world datasets relative to existing kernel search methods.

\section{Background}
\subsection{Gaussian Processes and the Role of Kernels}
Gaussian Processes (GPs) are a nonparametric Bayesian framework for modeling unknown mappings $f:\mathbb{R}^d \rightarrow \mathbb{R}$. Given input--output pairs $\mathcal{D}=\{(x_i, y_i)\}_{i=1}^n$, a GP prior $f\sim \mathcal{GP}(m,k)$ is fully specified by a mean function $m(x)$ and a positive--semidefinite covariance (kernel) function $k(x,x')$. Under a Gaussian likelihood, the posterior predictive distribution is available in closed form, and hyperparameters are learned by maximizing the log marginal likelihood (LML).

% DA: I suggest sticking with the same examples.  Before you used SE and now you are using RBF.
The kernel is the heart of the model: it defines similarity between inputs and governs smoothness, periodicity, and long-range dependence. Common base kernels include the squared-exponential (SE, also known as the radial basis function or RBF), Mat\'ern, rational quadratic (RQ), and periodic kernels. Complex behavior is often modeled by composing base kernels through addition or multiplication, for example
\[
  k(x,x') \;=\; k_{\mathrm{SE}}(x,x') + k_{\mathrm{PER}}(x,x'),
\]
to capture both smooth trends and seasonal oscillations. This composition view encodes rich inductive biases and underpins automated statistical modeling.

\subsection{Challenges in Kernel and Hyperparameter Optimization}
Despite its centrality, kernel selection remains a fundamental bottleneck. Even with a small grammar of base kernels and operators $\{+,\times\}$, the number of possible compositions grows exponentially with depth. A depth-3 grammar---combining three base kernels with two operators---can easily produce dozens to hundreds of valid structures.
% DA: Can you carefully define what you mean by depth-2 here and possible include a figure with an example?
Each kernel has continuous hyperparameters (length scales, periodicities, etc.) that strongly influence the LML; small changes can yield large shifts in predictive behavior, making naive grid search or gradient-based optimization unreliable. Additionally, every candidate kernel must be fitted and its LML evaluated, each requiring $\mathcal{O}(n^3)$ matrix operations for datasets of size $n$ \cite{williams2006gaussian}, which quickly becomes burdensome when hundreds of kernels are compared. Traditional practice---manually selecting kernels guided by intuition and trying a few alternatives---does not scale to modern, data-rich applications.

\subsection{Bayesian Optimization for Automatic Kernel Search and Kernel Distances}
Bayesian Optimization (BO) offers a principled way to automate kernel discovery. BO treats the LML as an expensive black-box function to be maximized. A surrogate model (i.e., another GP) is built over a search space of kernel structures, and an acquisition function such as Expected Improvement (EI) guides the selection of the next candidate. However, BO typically assumes a continuous, Euclidean input space. This assumption is problematic because kernel structures do not live in a natural Euclidean space: they are discrete objects generated by a grammar. Direct symbolic encodings therefore fail to yield meaningful geometric distances between kernels, making gradient-free BO difficult to apply directly.

This motivates defining kernel--kernel distances that render model space amenable to BO.

% DA:  These two subsections (the one above and below this comment, should be the same subsection.
% DA: I am concerned about the use of Hellinger distance given its tendency to 1.  Maybe something like symmetric KL divergence would be better (with a fixed number of sample locations or scaled by 1/N).  Another option is the Jensen-Bregman-LogDet Divergence though I'm not as familiar with that.

% SM: Experimented with using both KL divergence and Jensen–Bregman LogDet Divergence (instead of Hellinger as Dr. Allaire asked to try) for the pairwise kernel distance calculations.
% However, unlike Hellinger, both are non-Euclidean because they measure divergence on curved statistical manifolds, not straight-line distances in flat space. This curvature manifests as negative eigenvalues during MDS, which makes Euclidean embedding impossible. As a result, the feature representation required for BO through MDS featurization doesn’t work for KL or Jensen–Bregman LogDet Divergence. This is why our current implementation relies on Hellinger.

Recent research \cite{malkomes2016bayesian} has proposed measuring the probabilistic dissimilarity between kernels themselves. The core idea is to evaluate how differently two kernels induce distributions over functions. A particularly suitable metric is the Hellinger distance between two GP priors evaluated on a finite set of input locations. For kernels $k_i$ and $k_j$ with hyperparameters $\theta_i$ and $\theta_j$,
\[
  H^2\!\big(p(y\mid k_i,\theta_i),\, p(y\mid k_j,\theta_j)\big)
\]
quantifies the divergence between their induced output distributions. By marginalizing over hyperparameters using quasi-Monte Carlo (QMC) sampling over uniform bounds, we can obtain the expected squared Hellinger distance,
\[
  D_{ij} \;=\; \mathbb{E}_{\theta_i,\theta_j}\big[ H^2(\cdot) \big],
\]
which is robust to local hyperparameter variations and captures both structural and parameter uncertainty. The full matrix $D\in\mathbb{R}^{N\times N}$, computed for all kernel pairs, is symmetric with zeros on the diagonal and defines a kernel-kernel geometry.

Despite its convenience, the Hellinger-based dissimilarity matrix also has practical limitations. In many cases, the resulting distances become highly concentrated, i.e., most kernel pairs end up with values clustered near a common scale (often close to 1), which leads to a flat or low-contrast geometry. This concentration is expected because Hellinger distance is bounded in $[0,1]$ and, in high-dimensional Gaussian settings, even small changes in covariance structure can push distributions far apart; with hyperparameter marginalization, most kernel pairs saturate near the upper bound (concentration of measure). This makes subsequent tasks such as MDS embedding and Bayesian Optimization difficult, because the induced kernel-of-kernels landscape contains very little structural variation for BO to exploit. 

Beyond the Hellinger distance, other divergence measures such as KL divergence, Jensen–Shannon (JS) divergence, and transformed variants can also be used to construct kernel--kernel dissimilarity matrices. These metrics capture different aspects of probabilistic behavior: KL encodes directional information loss, JS provides a symmetric and smoothed variant, and $\sqrt{\mathrm{JS}}$ is a true metric that admits Euclidean embedding. However, each divergence induces a distinct geometric structure; some yield curved, non-Euclidean spaces with significant negative eigenvalues, while others produce high-dimensional manifolds that are difficult to embed faithfully. Selecting a divergence therefore balances theoretical metricity with practical spread in distances, which directly affects BO exploration. Practical kernel search benefits from a well-spread, approximately Euclidean distance matrix, ensuring stable MDS embeddings and an informative landscape for Bayesian optimization. In our framework, we systematically evaluate these divergences and their transformed variants (e.g., log-warping) to identify those that provide the most usable geometry for kernel optimization.

\subsection{Embedding Kernel Geometry into Euclidean Space}
For BO to exploit this geometry, we require a continuous representation of kernels. Classical Multidimensional Scaling (MDS) provides a standard embedding. Given a dissimilarity matrix $D$ with entries $D_{ij}=d_{ij}^2$ (expected squared distances), MDS finds low-dimensional coordinates $\{z_i\}_{i=1}^N \subset \mathbb{R}^p$ such that
\[
  \|z_i - z_j\|_2 \;\approx\; D_{ij}^{1/2}
  \quad \text{for all } i,j.
\]
In other words, MDS places kernels in a low-dimensional Euclidean space so that pairwise distances between points approximate the original dissimilarities.
These coordinates are not meant to be interpreted in isolation; instead, each vector $z_i$ compactly encodes the pattern of similarities of kernel $k_i$ to all others. This embedding transforms a discrete set of kernels into a continuous representation in $\mathbb{R}^p$ where Euclidean distances can accurately reflect probabilistic differences.\\

The combination of a valid distance metric between kernels and its MDS embedding makes it possible to apply GP-based BO to the kernel-selection problem. The embedded coordinates serve as inputs to the surrogate GP, while the objective remains the LML of the candidate kernels. Acquisition functions can now exploit uncertainty in this continuous space to efficiently identify promising composite kernels. This probabilistic geometric perspective defines the foundation of our kernel-of-kernels optimization, which we build upon to present the full methodology, including variance-aware distance penalties.

% -----------------------------------------------------
% METHODOLOGY
% -----------------------------------------------------
\section{Methodology}

In this work, we transform a discrete library of compositional kernels into a continuous, geometry-aware space on which Bayesian optimization (BO) can operate smoothly. The process has four components: (i) constructing a structured kernel library using a compositional grammar, (ii) defining probabilistic divergence measures between GP priors associated with these kernels, (iii) transforming and embedding the resulting kernel--kernel distance matrix into Euclidean space using multidimensional scaling (MDS), and (iv) performing BO directly in this embedded space while evaluating only discrete kernel candidates. Each step reshapes a combinatorial model search problem into a continuous optimization problem with meaningful geometry.

Although the MDS embedding defines a continuous coordinate system, our search remains discrete (acquisition proposals are \emph{snapped} to the nearest embedded kernel).

\subsection{Grammar-Based Kernel Library}

We begin by defining a symbolic grammar that generates a large, expressive set of candidate kernels. This grammar is closed under addition and multiplication, ensuring that any subexpression may be expanded using a base kernel or combined multiplicatively with another. The base kernels themselves, such as squared-exponential, periodic, linear, and rational quadratic, are interchangeable under the grammar rules. Starting from a minimal seed set, recursive application of these rules produces increasingly complex kernel expressions, yielding a structured but finite kernel library. Each kernel in this library is treated as a distinct model class whose hyperparameters are marginalized over when computing its predictive distribution. Importantly, the symbolic form is not used directly for optimization; instead, it provides the initial discrete set of candidate kernels that will be analyzed and embedded into a continuous space.

\subsection{Divergence Between GP Priors}

To quantify similarity between kernels, we compare the probability distributions they induce over functions, rather than their symbolic structure or hyperparameter vectors. For each kernel $k_i$ in a discrete library, we consider the associated GP prior evaluated on a fixed reference set of inputs $X = \{x_s\}$. Conditioned on hyperparameters $\theta$, this prior induces a multivariate normal distribution over function values. Kernel comparison can therefore be reduced to the comparison of Gaussian measures.

We measure dissimilarity between kernels using probabilistic divergence measures defined between the induced GP priors. Given two kernels $k_i$ and $k_j$ with hyperparameter distributions (uniform over predefined bounds), we approximate their expected divergence by sampling hyperparameters via quasi--Monte Carlo integration and averaging the divergence between the resulting predictive distributions. This yields a hyperparameter-marginalized kernel--kernel dissimilarity that reflects both structural and parametric uncertainty.

Multiple probabilistic divergences are viable candidates for kernel comparison, each inducing a distinct geometric structure. Among these, the squared Hellinger distance provides a symmetric and bounded measure of separation between Gaussian measures~\cite{ding2023empirical} and has been used in model-space Bayesian optimization~\cite{malkomes2016bayesian}. However, as noted earlier, its boundedness implies that pairwise distances between GP priors may saturate near their upper limit, producing distance matrices with limited dynamic range. Characterizing how this behavior affects geometry and downstream optimization is necessary for constructing reliable kernel embeddings.

To compute distances between kernels in practice, we compare the Gaussian Process priors they induce on a fixed set of reference input locations. Although GP priors are infinite-dimensional objects, their restriction to a finite input set yields multivariate Gaussian distributions that can be compared using standard probabilistic divergences. Specifically, we discretize the input domain using a shared reference set of $N_{\mathrm{ref}} = 50$ uniformly spaced points and evaluate the GP prior covariance matrices induced by each kernel on this grid.

Distances between kernels are then computed by measuring the divergence between the corresponding finite-dimensional Gaussian priors defined on this common reference set. This construction ensures that all kernel comparisons are performed in a consistent functional space, independent of the dataset used for downstream optimization. The resulting pairwise distance matrix captures differences in the induced prior distributions rather than symbolic structure or hyperparameter values alone.

The distance matrix is computed once prior to Bayesian Optimization and subsequently embedded into a Euclidean space using multidimensional scaling. The resulting coordinates serve as kernel descriptors and define the input space over which Bayesian Optimization is performed.

We next analyze the Euclidean embeddability of these distances and how to correct curvature when needed.

\subsection{Euclidean Geometry and Curvature Analysis}

Bayesian optimization over kernel space requires a representation in which distances admit a faithful Euclidean embedding. Classical multidimensional scaling (MDS) provides such an embedding when the underlying distance matrix is Euclidean~\cite{zhang2010multidimensional}. A symmetric matrix of pairwise distances is Euclidean if and only if the corresponding double-centered Gram matrix is positive semidefinite. Computing the eigenvalues of this Gram matrix therefore provides a diagnostic of the curvature of the underlying geometry. For N choose 2 pairwise distances, if all eigenvalues of the double centered Gram Matrix of the distance matrix are nonnegative, the distance matrix corresponds to distances between points in some Euclidean space of dimension at most $N-1$. 

On the other hand, negative eigenvalues in the double-centered Gram matrix signal that the distances arise from a curved (non-flat) manifold, such as spherical or hyperbolic geometry, and therefore cannot be embedded isometrically into Euclidean space. In this setting, divergences such as Kullback--Leibler (KL) and Jensen--Shannon (JS) frequently produce distance matrices with substantial negative eigenvalues, reflecting the curved statistical manifolds on which these divergences are naturally defined. Hellinger-based distances may also exhibit curvature when hyperparameter variability induces nonlinear distortions, compounding boundedness effects. Such curvature degrades the fidelity of Euclidean embeddings and complicates acquisition optimization in Bayesian optimization.

\subsection{Transformations for Euclidean Embeddability}

When the kernel-kernel distance matrix exhibits curvature, as indicated by negative eigenvalues in the double-centered Gram matrix, the underlying geometry is non-Euclidean and cannot be embedded faithfully using classical MDS. In such cases, the distance structure must be corrected or transformed before embedding. Two classes of transformations are particularly effective for reducing curvature and yielding distances that are closer to Euclidean in practice.

\begin{figure}[h]
    \centering
    \includegraphics[width=0.4\textwidth]{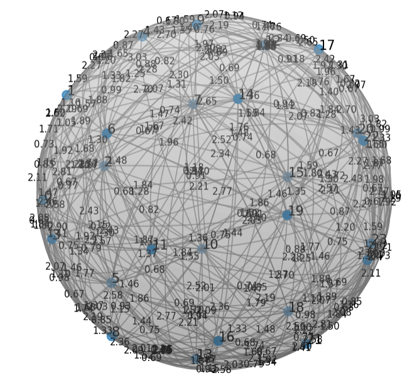}
    \caption{Geodesic (great-circle) distances between points on a sphere. These distances arise from a curved Riemannian manifold and cannot be embedded exactly into Euclidean space.}
    \label{fig:sphere_geodesic}
\end{figure}

A central example arises when the dissimilarities between kernels behave like geodesic distances on a curved manifold, such as a sphere. In this scenario, the distances correspond to great-circle lengths, which fundamentally reside on a non-flat Riemannian manifold. MDS applied directly to such distances cannot recover a faithful embedding, even with arbitrarily high embedding dimension, because no point configuration in Euclidean space reproduces all pairwise geodesic distances exactly. This effect manifests in practice as non-vanishing reconstruction error and persistent negative eigenvalues in the Gram matrix (Fig.~\ref{fig:sphere_geodesic}).

\begin{figure}[h]
    \includegraphics[width=0.5\textwidth]{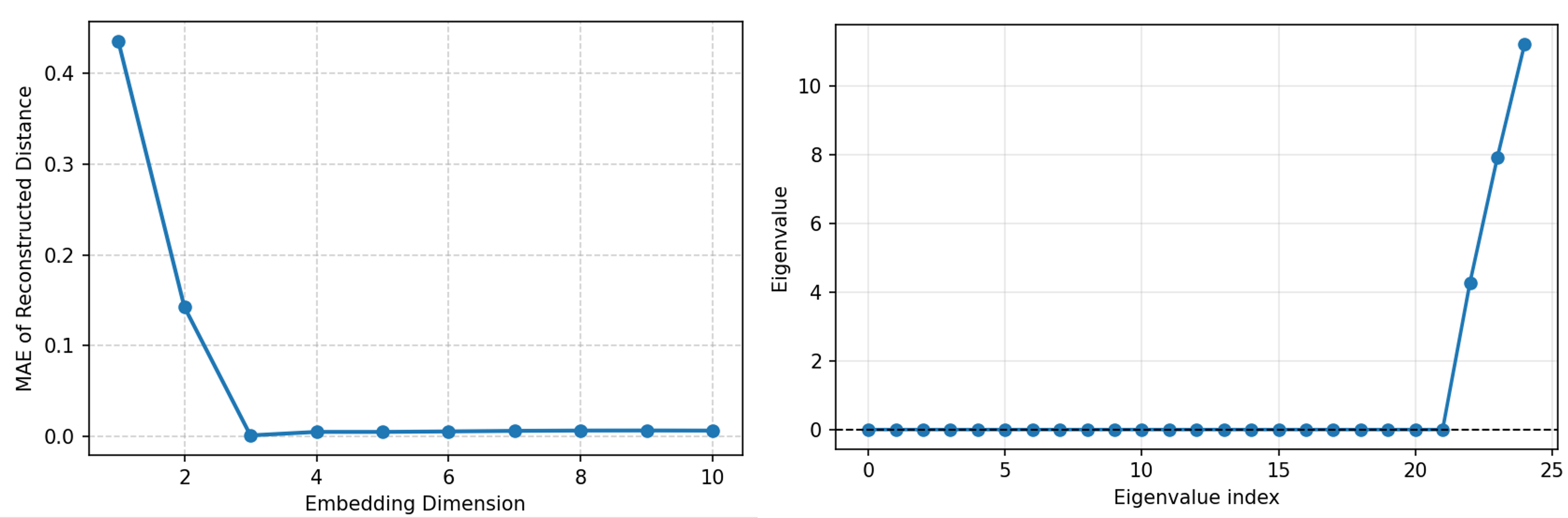}
    \caption{Chordal-mapping correction for spherical geometry. Left: reconstruction error from MDS collapses to zero at $k=3$. Right: eigenvalue spectrum shows elimination of negative eigenvalues after the transformation, confirming Euclidean embeddability.}
    \label{fig:chordal_results}
\end{figure}

To address this, we employ the Euclidean chordal mapping, which converts spherical geodesic distances into Euclidean chord lengths. If two points on a unit sphere are separated by a geodesic distance $\delta$, their chordal distance is $2 \sin(\delta / 2)$. This transformation maps geodesic distances to chord lengths, yielding a flat Euclidean geometry for the spherical case while preserving distance ordering. After applying the chordal mapping, the distance matrix is Euclidean (up to numerical precision) for this spherical case, as evidenced by a Gram matrix with no negative eigenvalues (Fig.~\ref{fig:chordal_results}). The MDS reconstruction error drops to zero in three dimensions—the intrinsic dimension of the embedding—showing that the curved manifold has been flattened. This example illustrates how curvature in probabilistic distance matrices can obstruct embedding and how geometric mappings can restore Euclidean structure.

\begin{figure}[h]
    \centering
    \includegraphics[width=0.4\textwidth]{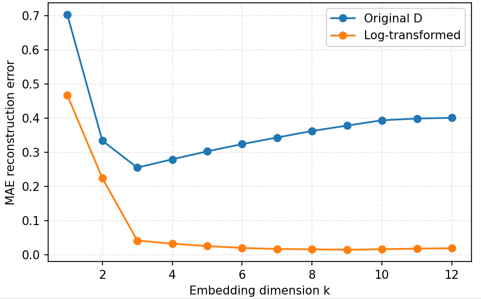}
    \caption{Comparison of MDS reconstruction error before and after log-warping. The log-transformed distances produce significantly lower error and a smoother Euclidean embedding.}
    \label{fig:logwarp_mae}
\end{figure}

\begin{figure}[h]
    \includegraphics[width=0.5\textwidth]{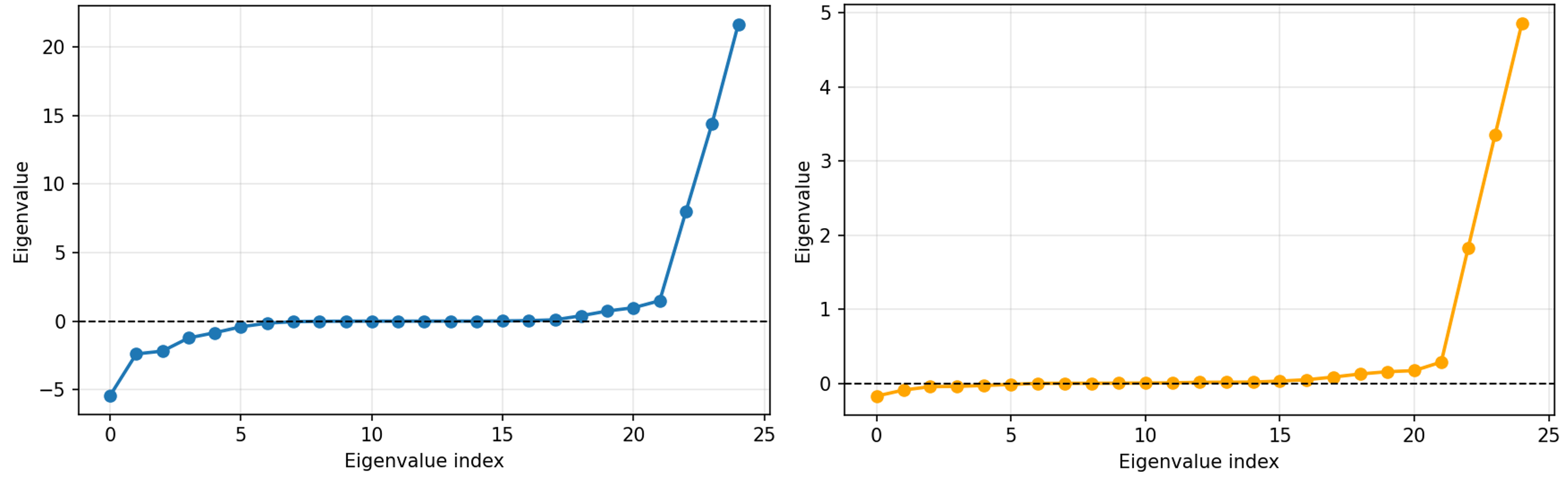}
    \caption{Eigenvalue spectra of the double-centered Gram matrix before (left) and after (right) log-warping. Log-warping removes large negative eigenvalues, substantially reducing curvature.}
    \label{fig:logwarp_eigs}
\end{figure}

Another simple transformation is logarithmic warping, which is broadly applicable beyond spherical geometry. Divergences between GP priors often exhibit long-tailed or highly compressed distributions, where large distances dominate and small distances cluster tightly. When raw distances vary across several orders of magnitude or collapse near a single value—as is frequently the case for Hellinger and KL divergences—the curvature of the induced geometry increases, and the Gram matrix acquires negative eigenvalues of nontrivial magnitude. Log-warping the distances counteracts this behavior by compressing large distances more strongly than small ones, redistributing variance and smoothing curvature. After the transformation $d' = \log(d + \epsilon)$, the Gram matrix typically retains very few negative eigenvalues, and the remaining ones are negligible in magnitude (Figs.~\ref{fig:logwarp_mae} and \ref{fig:logwarp_eigs}). This spectral improvement indicates that the transformed distances are closer to a flat Euclidean manifold and embed with lower distortion under MDS.

The spherical chordal mapping and logarithmic warping exemplify how geometric distortions in kernel distance matrices can be detected and corrected. The first treats a canonical form of curvature arising from manifold structure, while the second addresses statistical curvature arising from divergence concentration. By selecting the transformation that best reduces curvature for the divergence at hand, we aim to obtain a stable, low-distortion Euclidean embedding. This step is essential for obtaining reliable MDS coordinates and, consequently, a well-behaved landscape for Bayesian Optimization on the kernel manifold.

% ---------------------------------------------------------
% Figure 1: Geodesic distances on sphere
% ---------------------------------------------------------

% ---------------------------------------------------------
% Figure 2: Chordal mapping (MAE + eigenvalues)
% ---------------------------------------------------------

% ---------------------------------------------------------
% Figure 3: Log-warping MAE
% ---------------------------------------------------------

% ---------------------------------------------------------
% Figure 4: Log-warping eigenvalue corrections
% ---------------------------------------------------------

Before selecting a divergence to serve as the basis of our kernel-of-kernels geometry, it is necessary to understand how various candidates behave with respect to Euclidean embeddability. KL divergence, Hellinger distance, and Jensen--Shannon divergence each produce different spectral and geometric distortions in the distance matrix, often necessitating transformations to achieve a usable Euclidean structure. The following discussion characterizes these divergences and motivates the choice of a transformation that yields a stable MDS embedding for Bayesian optimization.

\paragraph{Squared Hellinger distance.}
The squared Hellinger distance is defined as
\begin{equation}
  H^2(p,q)
  = \frac{1}{2} \int \bigl(\sqrt{p(x)} - \sqrt{q(x)}\bigr)^2 \, dx.
\end{equation}
Hellinger distance is a true metric: it is symmetric, non-negative,
and satisfies the triangle inequality. Moreover,
$H(p,q)$ corresponds to the Euclidean distance between the
\emph{square-root densities} $\sqrt{p}$ and $\sqrt{q}$ in the $L^2$ space.
This linearization---mapping $p \mapsto \sqrt{p}$---implies that the underlying geometry
lies on the unit sphere in $L^2$. Specifically,
\begin{equation}
  H^2(p,q) = 1 - \int \sqrt{p(x)\,q(x)} \, dx = 1 - \langle \sqrt{p}, \sqrt{q} \rangle,
\end{equation}
so that $\langle \sqrt{p}, \sqrt{q} \rangle$ acts as a cosine similarity
between points on the unit sphere in $L^2$. This spherical embedding contributes to
stability and symmetry, making Hellinger distance appealing for kernel-of-kernels
geometry.
% === INSERT FIGURE SET 1 HERE ===
\begin{figure}[h]
    \includegraphics[width=0.5\textwidth]{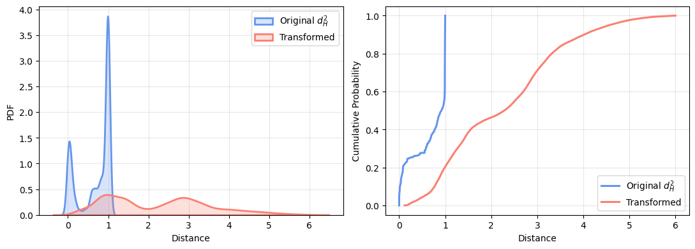}
    \caption{
    Probability density (left) and cumulative distribution (right) of the original squared Hellinger distances (blue) between GP priors compared with their log-warped counterparts (red). The raw distances are highly compressed near~1, producing a nearly flat geometry that is difficult for MDS to embed. Log-warping expands the dynamic range, increases contrast, and produces a more informative and usable distance distribution for Euclidean embedding and Bayesian Optimization.
    }

    \label{fig:hellinger_transform_pdfcdf}
\end{figure}

% === FIGURE: High-Dimensional MDS Reconstruction Error ===
\begin{figure}[h]
    \centering
    \includegraphics[width=0.4\textwidth]{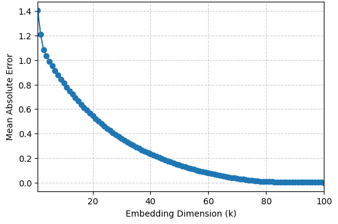}
    \caption{MDS reconstruction error as a function of embedding dimension for the
    log-transformed Hellinger distance matrix. Although log-warping improves
    curvature and stabilizes the Gram spectrum, the induced geometry remains
    high-dimensional: the reconstruction error decays slowly and only approaches
    zero at large embedding dimensions ($k \approx 80$--$100$), indicating that
    the transformed distance matrix has high intrinsic rank.}
    \label{fig:mds_highdim_recon}
\end{figure}

% === INSERT FIGURE SET 2 HERE ===
\begin{figure}[t]
    \includegraphics[width=0.5\textwidth]{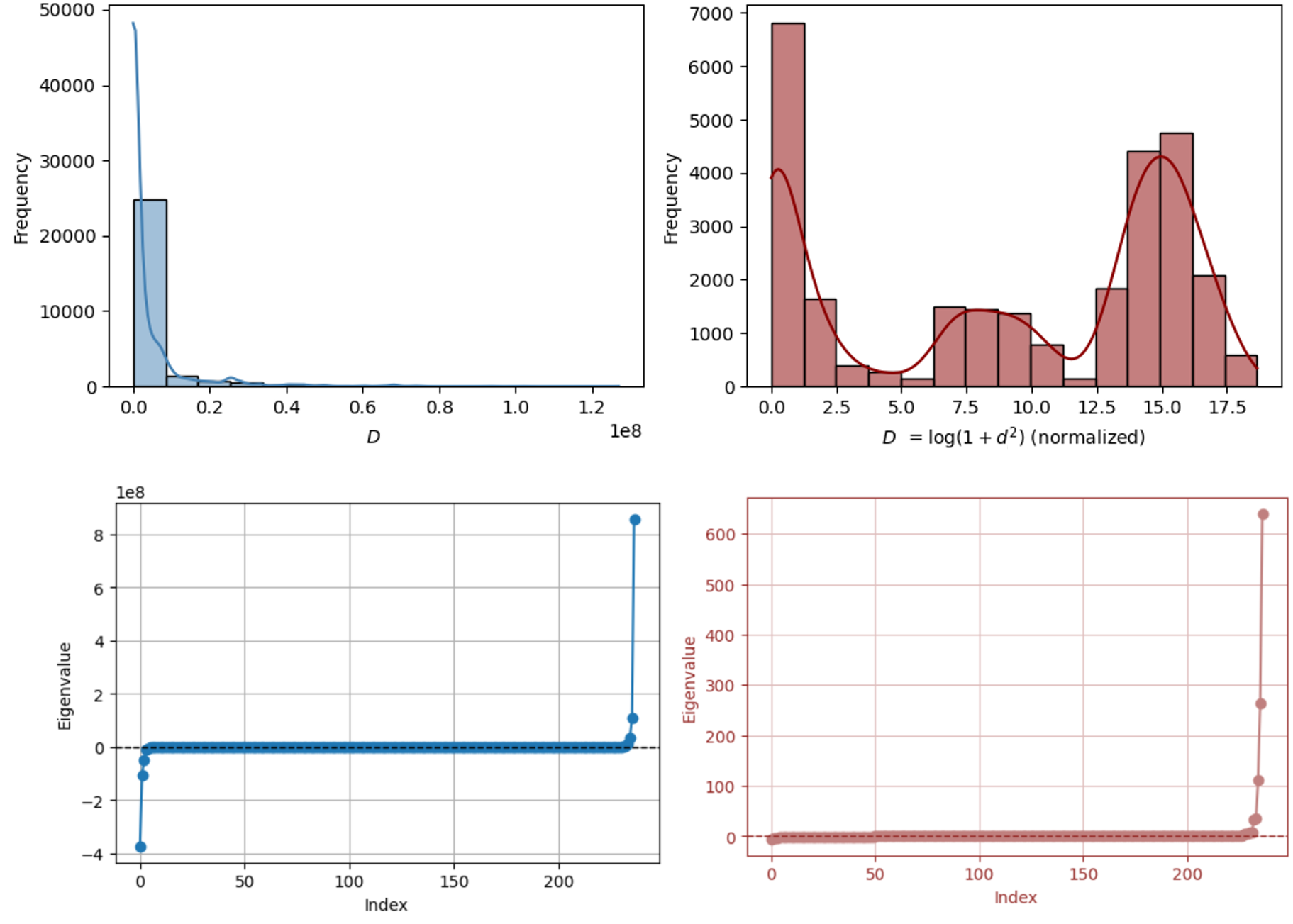}
    \caption{
Effect of logarithmic warping on KL-based kernel distances. Top row: the original $D=d_{\mathrm{KL}}^2$ distances (left) are extremely heavy-tailed and statistically curved, while the warped distances $D = \log(1 + d_{\mathrm{KL}}^2)$ (right) become well-spread and approximately Gaussian. Bottom row: eigenvalue spectra of the double-centered Gram matrices before (left) and after (right) transformation. The raw KL distances exhibit strong non-Euclidean curvature with many large-magnitude negative eigenvalues; after warping, only a few small-magnitude negative eigenvalues remain, indicating an approximately Euclidean geometry that can be faithfully embedded via MDS.
}

    \label{fig:KL_transform_eigs}
\end{figure}

However, spherical geometry also comes with a drawback:
when the square-root densities of many kernels lie close to one another,
distances compress on the sphere.
In practice, this produces the well-known \emph{distance concentration} effect,
where most $H^2(p,q)$ values cluster close to 1.
This collapse reduces geometric contrast and harms MDS embeddings,
flattening the induced manifold into a nearly uniform cloud
(see Fig.~\ref{fig:hellinger_transform_pdfcdf}).
For this reason, we apply monotone transformations such as 
$D = \log\!\left(1 + H^2\right)$ to expand the dynamic range
while preserving ordering, and improve embeddability.

For the Hellinger divergence, log-warping is applied primarily to stretch the heavy concentration of distances near 1, creating a more informative spread for optimization. However, this increased spread introduces additional geometric variation, which in turn increases the intrinsic dimensionality required for an accurate MDS embedding. As a result, the log-transformed Hellinger matrix often embeds only in relatively high dimensions (Fig.~\ref{fig:mds_highdim_recon}), even though the warping significantly improves the BO landscape.

\paragraph{Kullback--Leibler divergence.}
The Kullback--Leibler (KL) divergence between two probability densities 
$p(x)$ and $q(x)$ is defined as
\begin{equation}
    Div_{\mathrm{KL}}(p \,\|\, q)
    = \int p(x) \log \frac{p(x)}{q(x)} \, dx.
\end{equation}
KL measures the expected information loss when $q$ is used to approximate $p$ \cite{ji2020kullback};
informally, it quantifies how many \emph{extra} nats (or bits) are needed to encode
samples from $p$ when a code optimized for $q$ is used.
A key property of KL divergence is its asymmetry:
\begin{equation}
  Div_{\mathrm{KL}}(p \,\|\, q) \neq Div_{\mathrm{KL}}(q \,\|\, p),
\end{equation}
except in the special case $p=q$ almost everywhere.

From an information-geometric perspective, $D_{\mathrm{KL}}$ is the canonical
Bregman divergence generated by the negative Shannon entropy,
and therefore lives on a \emph{curved statistical manifold}.
This has an important implication for kernel geometry:
KL divergence is \emph{not} a metric \cite{ji2020kullback}. It violates symmetry and the triangle inequality, and
its induced geometry is non-Euclidean.
As a consequence, pairwise KL distance matrices often exhibit
large negative eigenvalues, making them unsuitable as distances for classical MDS
without additional transformations or regularization.
Empirically, KL divergences computed between Gaussian Process priors tend to be extremely
skewed, with a heavy tail and most values concentrated near zero (see Fig.~\ref{fig:KL_transform_eigs}).
This motivates using symmetric or Euclideanized variants of KL, or applying monotone warping
transformations such as logarithmic scaling.

\paragraph{$\sqrt{\mathrm{JS}}$ Divergence (as a True Metric)}
The square-root Jensen–Shannon (JS) divergence serves as an alternative to Hellinger distance because it is also a true metric on the space of probability distributions. Whereas the raw JS divergence is symmetric and bounded but does not itself satisfy the triangle inequality, the square-root form, $\sqrt{\mathrm{JS}(p,q)}$, does satisfy all metric axioms, as rigorously established by Endres and Schindelin \cite{endres2003new}. This metricity implies that the induced distance matrix is geometrically well-behaved: it corresponds to a negative-type kernel and therefore admits an exact isometric embedding into a Hilbert space. In contrast to divergences such as KL or others, whose induced distance matrices often exhibit curvature, negative eigenvalues in the double-centered Gram matrix, and poor low-dimensional reconstruction, the sqrt JS distance tends to produce significantly flatter geometries with fewer violations of Euclidean structure. The boundedness of sqrt JS (between 0 and $\log 2$) also avoids the extreme compression seen in Hellinger or the heavy-tailed behavior characteristic of KL, yielding distances that are better distributed across scales. As a consequence, MDS embeddings based on this metric typically show improved spectral stability, reduced curvature, and more faithful low-dimensional structure, making it a strong candidate for kernels geometry and for downstream Bayesian Optimization on the embedded manifold.

\begin{figure}[h]
    \centering
    \includegraphics[width=0.43\textwidth]{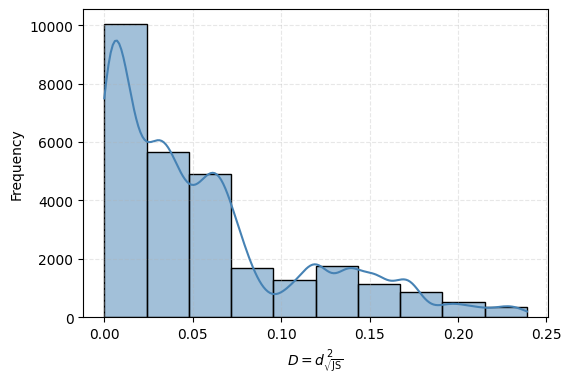}
    \caption{Empirical distribution of $D=d_{\sqrt{JS}}^2$ between GP priors. The distribution is bounded and well-behaved, with moderate spread, reflecting the metric nature and stability of the sqrt Jensen--Shannon divergence.}
    \label{fig:js_hist}
\end{figure}

\begin{figure}[h]
    \centering
    \includegraphics[width=0.4\textwidth]{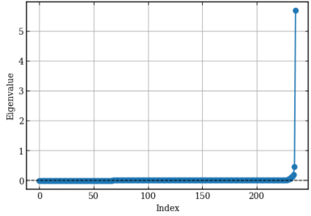}
    \caption{Eigenvalue spectrum of the double-centered Gram matrix constructed from the squared distance matrix $D=d_{\sqrt{JS}}^2$. All eigenvalues are non-negative, confirming exact Euclidean embeddability and the absence of curvature.}
    \label{fig:js_eig}
\end{figure}

To validate the Euclidean properties of the $\sqrt{\mathrm{JS}}$ divergence in practice, we computed the full pairwise squared distance matrix using $\sqrt{\mathrm{JS}}$ as the distance metric between all GP priors in our kernel library and analyzed its spectral and geometric structure. The empirical distribution of the squared distances are shown in Fig.~\ref{fig:js_hist}, revealing a highly concentrated but strictly positive spread, consistent with the theoretical boundedness of sqrt Jensen-Shannon divergence in $[0, \log 2]$. More importantly, the eigenvalue spectrum of the double-centered Gram matrix (Fig.~\ref{fig:js_eig}) confirms that \emph{all eigenvalues are non-negative}, up to numerical precision. This verifies that the double-centered Gram matrix is positive semidefinite, so the distance matrix is Euclidean-embeddable (up to numerical precision). In contrast to KL and Hellinger distances, no curvature-correction or warping transformation is required for $\sqrt{\mathrm{JS}}$, making it a robust default metric for kernel geometry.

% === FIGURE: JS HISTOGRAM ===

% === FIGURE: JS EIGENVALUE SPECTRUM ===

\subsection{Multidimensional Scaling of Kernel Geometry}

With a Euclidean squared distance matrix ($D=d_{\sqrt{JS}}^2$) in hand, we embed the kernels using classical multidimensional scaling. MDS constructs an embedding by double-centering the squared distances to form a Gram matrix, performing eigendecomposition, and computing coordinates from the positive eigenvalues and their corresponding eigenvectors. The resulting coordinates provide a continuous representation in which Euclidean distances mirror the probabilistic dissimilarities between kernels.

% === FIGURE: Eigenvalue decay / embedding quality ===
\begin{figure}[h]
    \centering
    \includegraphics[width=0.47\textwidth]{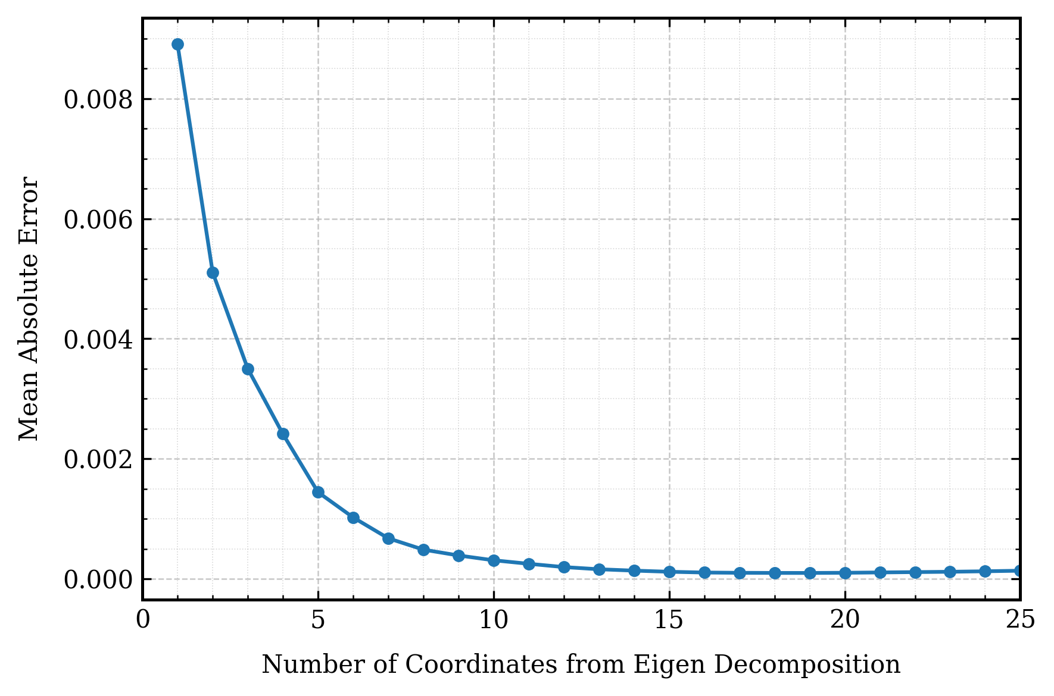}
    \caption{Mean absolute reconstruction error as a function of the number of MDS coordinates retained. The rapid eigenvalue decay indicates that the kernel geometry is effectively low-dimensional.}
    \label{fig:mds_coord_error}
\end{figure}

These coordinates are not intended to be interpretable individually; their meaning arises from the geometry they preserve. Kernels that induce similar GP priors lie close to each other, while those producing distinct distributions lie farther apart. MDS thus transforms a structured but discrete kernel library into a continuous representation in $\mathbb{R}^p$ that reflects the intrinsic structure of the kernel space. Importantly, the embedding often lives in a low-dimensional space (e.g., $10$--$20$ dimensions), even when the original kernel library is large, due to strong geometric correlations among compositional kernels.

\begin{figure}[h]
    \centering
    \includegraphics[width=0.4\textwidth]{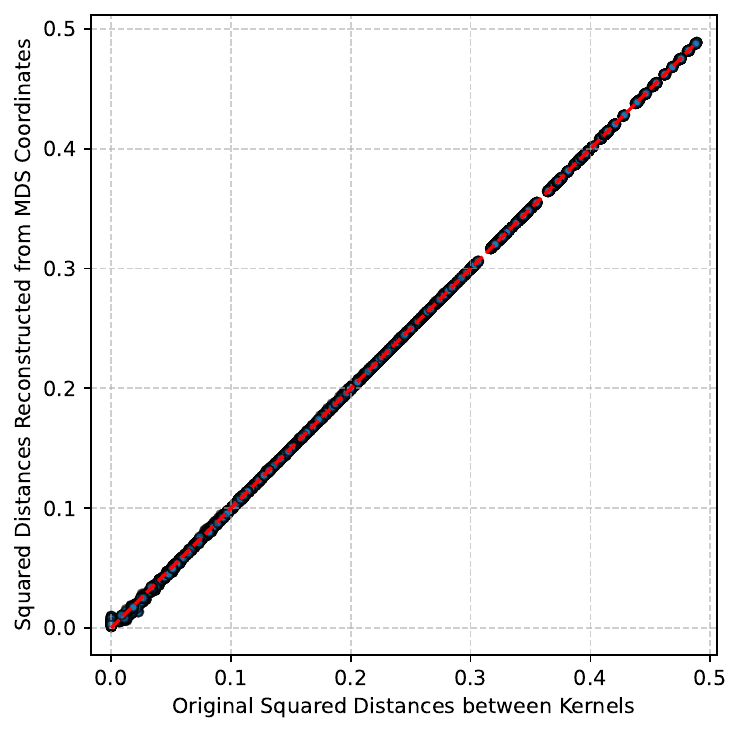}
    \caption{Reconstructed distances from the MDS embedding versus the true (Hellinger-based) kernel distances. The near-perfect alignment indicates high fidelity of the Euclidean embedding.}
    \label{fig:mds_reconstruction}
\end{figure}

Fig.~\ref{fig:mds_coord_error} and Fig.~\ref{fig:mds_reconstruction} show that the kernel coordinates in the MDS space well represent the true pairwise distances.

% === FIGURE: Distance reconstruction vs true distances ===

\begin{figure}[h]
    \centering
    \includegraphics[width=0.45\textwidth]{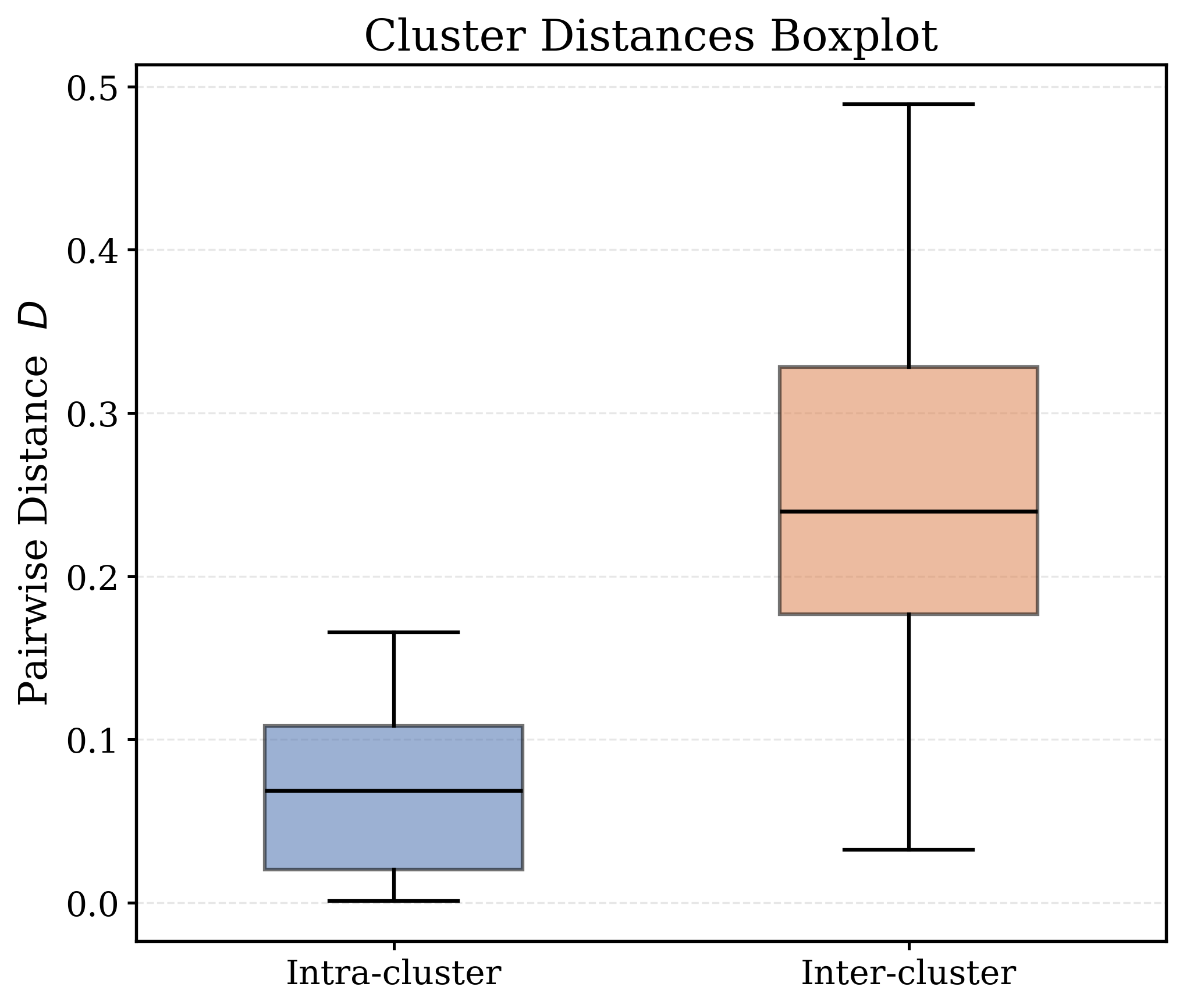}
    \caption{
    Boxplot comparison of intra-cluster and inter-cluster pairwise distances obtained from
    $k$-means clustering ($k=5$) in the 15-dimensional kernel embedding.
    Intra-cluster distances remain tightly concentrated near zero, while inter-cluster distances
    span a substantially broader range, demonstrating clear geometric separation between clusters.
    }
    \label{fig:cluster_boxplot}
\end{figure}

\begin{figure}[h]
    \centering
    \includegraphics[width=0.45\textwidth]{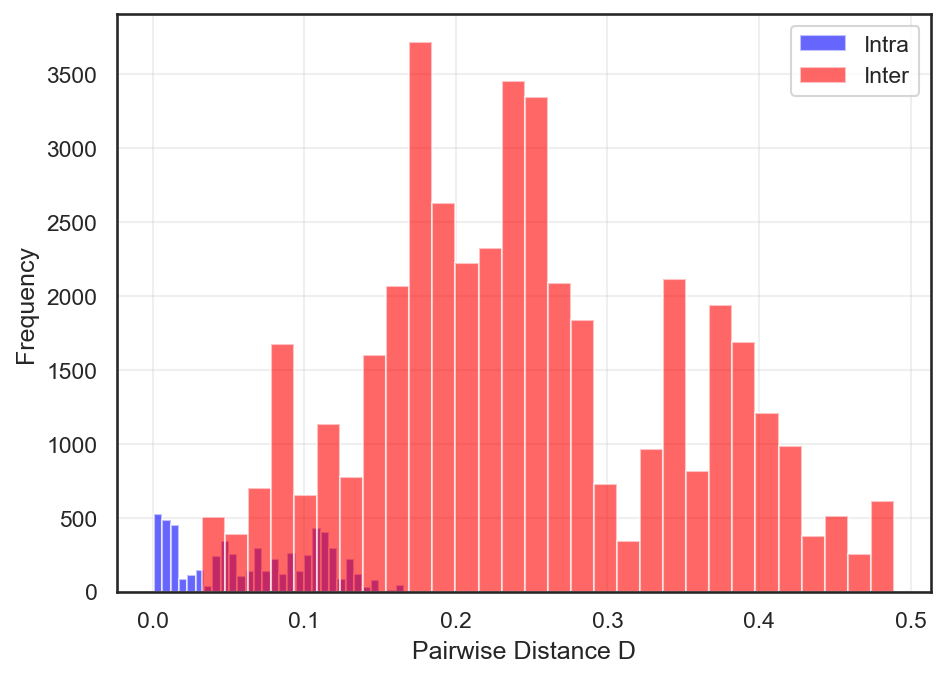}
    \caption{
    Histogram of pairwise distances for intra-cluster and inter-cluster kernel pairs.
    Intra-cluster distances remain narrowly concentrated near zero ($D < 0.1$),
    whereas inter-cluster distances cover a much wider range ($0.1 < D < 0.5$),
    confirming that the embedding preserves meaningful structural similarity.
    }
    \label{fig:cluster_hist}
\end{figure}

The validity of the proposed kernel manifold is further assessed through independent structural analyses performed directly in the embedded space. Using $k$-means clustering ($k=5$) on the 15-dimensional MDS coordinates, we observe a pronounced separation between intra-cluster and inter-cluster pairwise distances when measured with the proposed probabilistic metric. As shown in Fig.~\ref{fig:cluster_boxplot} and Fig.~\ref{fig:cluster_hist}, intra-cluster distances remain tightly concentrated near zero (dissimilarity $D<0.1$), whereas inter-cluster distances span a substantially broader range ($0.1<D<0.5$). This clear separation indicates that kernels grouped together in the embedding induce highly similar Gaussian-process priors, while dissimilar kernels are well separated geometrically.

\begin{figure}[h]
    \centering
    \includegraphics[width=0.48\textwidth]{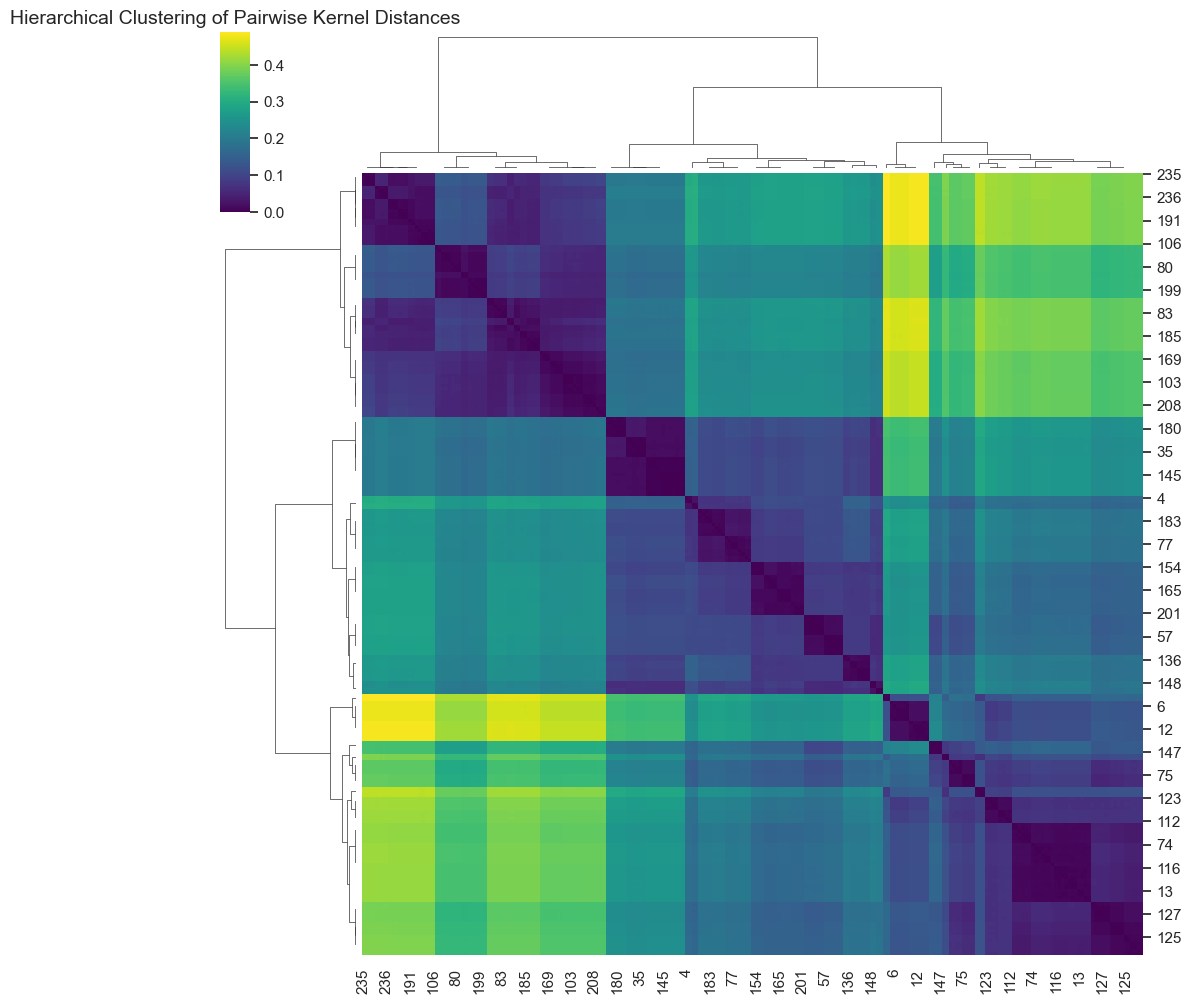}
    \caption{
    Hierarchical clustering of pairwise kernel distances derived from the proposed probabilistic metric.
    The dendrogram shows that closely related kernels merge at low linkage distances, while increasingly
    dissimilar kernels coalesce only at larger distances. Reordering the distance matrix according to
    this hierarchy reveals a clear block-diagonal structure, indicating coherent clusters of
    probabilistically similar kernels.
    }
    \label{fig:hierarchical_clustering}
\end{figure}

Complementary hierarchical clustering provides further confirmation of this structure. The dendrogram in Fig.~\ref{fig:hierarchical_clustering} shows that closely related kernels merge at very low linkage distances, while increasingly dissimilar kernels coalesce only at larger distances. When the pairwise distance matrix is reordered according to the hierarchical structure, a pronounced block-diagonal pattern emerges, revealing regions of consistently small intra-group distances. These analyses demonstrate that the 15-dimensional embedding preserves meaningful structural similarity: kernels that are close in the embedded space remain close under both $k$-means and hierarchical clustering, providing a faithful geometric substrate for Bayesian Optimization to reason about similarity, diversity, and exploration.

Although the manifold is continuous, the kernel library remains discrete. BO operates over the continuous embedding, but when the acquisition function proposes a new point, we \emph{snap} it to the nearest embedded kernel. This avoids inverting the embedding or generating new kernels analytically, while still allowing BO to benefit from continuous spatial reasoning.

\subsection{Surrogate Model for Kernel Space (Kernel-of-Kernels)}

To apply Bayesian Optimization over the kernel space, we construct a Gaussian Process surrogate whose inputs are kernel descriptors rather than physical inputs. Each kernel in the library is represented by a point $z_i \in \mathbb{R}^p$ obtained from the MDS embedding of the divergence-based kernel--kernel distance matrix. The objective associated with each point is the log marginal likelihood of the corresponding GP model fitted to the data.

A covariance function must therefore be defined directly over the kernel space. In our framework, this is achieved by endowing the embedded kernel coordinates with a geometry-aware kernel-of-kernels covariance. We begin with a stationary radial basis function (RBF) kernel on the MDS coordinates,
\[
k_{\mathcal{K}}(z_i, z_j)
\;=\;
\sigma^2 \exp\!\left(-\frac{\|z_i - z_j\|_2^2}{2\ell^2}\right),
\]
where the Euclidean distances $\|z_i - z_j\|_2$ approximate the transformed probabilistic dissimilarities between the corresponding GP priors. Because the embedding preserves kernel--kernel distances, this covariance assigns high similarity to kernels inducing similar distributions and low similarity to kernels with distinct inductive biases.

In experiments, we also evaluate a multi-scale kernel-of-kernels surrogate modeled as a mixture of RBF components with different length scales,
\begin{equation}
\begin{aligned}
k_{\mathcal{K}}^{\text{multi}}(z_i, z_j)
&= \sum_{m=1}^{M} w_m \exp\!\left(-\frac{\|z_i - z_j\|_2^2}{2\ell_m^2}\right), \\
&\qquad w_m \ge 0,\;\sum_{m=1}^{M} w_m = 1.
\end{aligned}
\end{equation}
which captures both local and global structure in the kernel manifold. This construction yields a Gaussian Process defined over the space of kernels itself, allowing BO to exploit smoothness and uncertainty structure in kernel space. Importantly, the geometry encoded in the MDS embedding ensures that neighborhoods in this space correspond to meaningful functional similarity, rather than arbitrary symbolic proximity.

When a multi-scale surrogate is used, the kernel-of-kernels covariance is modeled as a weighted sum of RBF components with different length scales, whose weights and length scales are learned by maximizing the marginal likelihood. In this sense, essentially the surrogate itself defines a kernel over the kernels-of-kernels manifold.

As baselines, we compare against random selection from the discrete kernel library and an LLM-guided kernel search strategy (described in the Results section).

\subsection{Bayesian Optimization on the Kernel Manifold}
The final step is to perform BO directly over the MDS-embedded space. A Gaussian Process surrogate is placed over the embedded coordinates, using a standard kernel such as squared-exponential or Matérn to capture smooth variation in model evidence across the manifold. The objective function remains the log marginal likelihood of the GP model defined by a candidate kernel. Because this evaluation is expensive, it requires hyperparameter learning and posterior computations, the BO framework provides a sample-efficient mechanism for identifying promising kernels.

The acquisition function, typically Expected Improvement, is optimized over the continuous latent space. Acquisition maximization yields a continuous coordinate that is then matched to the nearest embedded kernel for evaluation. The surrogate model is updated with this new observation, and the process repeats. Over iterations, BO explores the kernel manifold in a principled manner, preferring regions with high uncertainty or high predicted evidence while avoiding redundant evaluations of similar kernels.

% -----------------------------------------------------
% EXPERIMENTS
% -----------------------------------------------------
\section{Results and Discussion}

\begin{figure*}[h!]
    \centering
    \begin{tabular}{ccc}
        \includegraphics[width=0.30\textwidth]{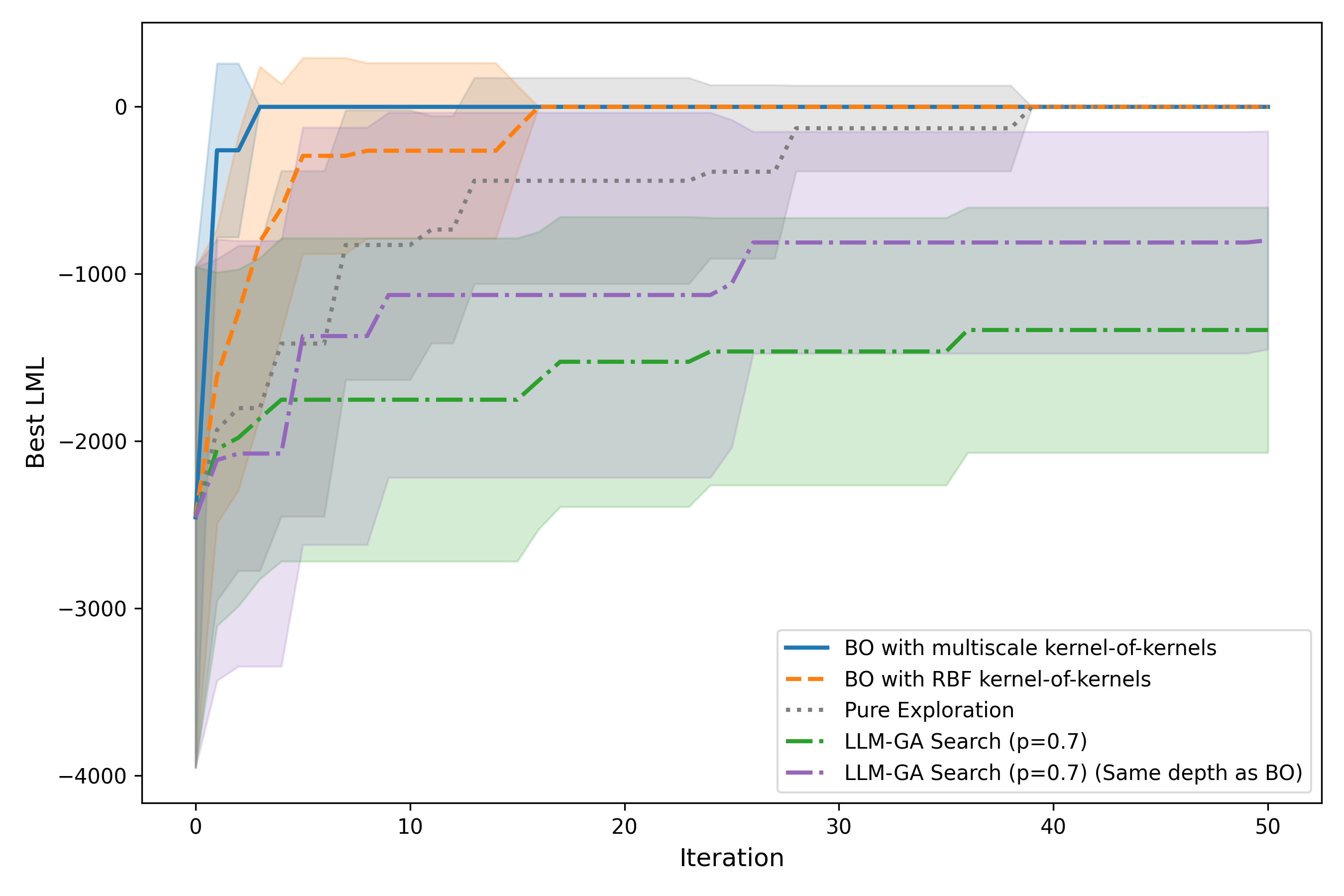} &
        \includegraphics[width=0.30\textwidth]{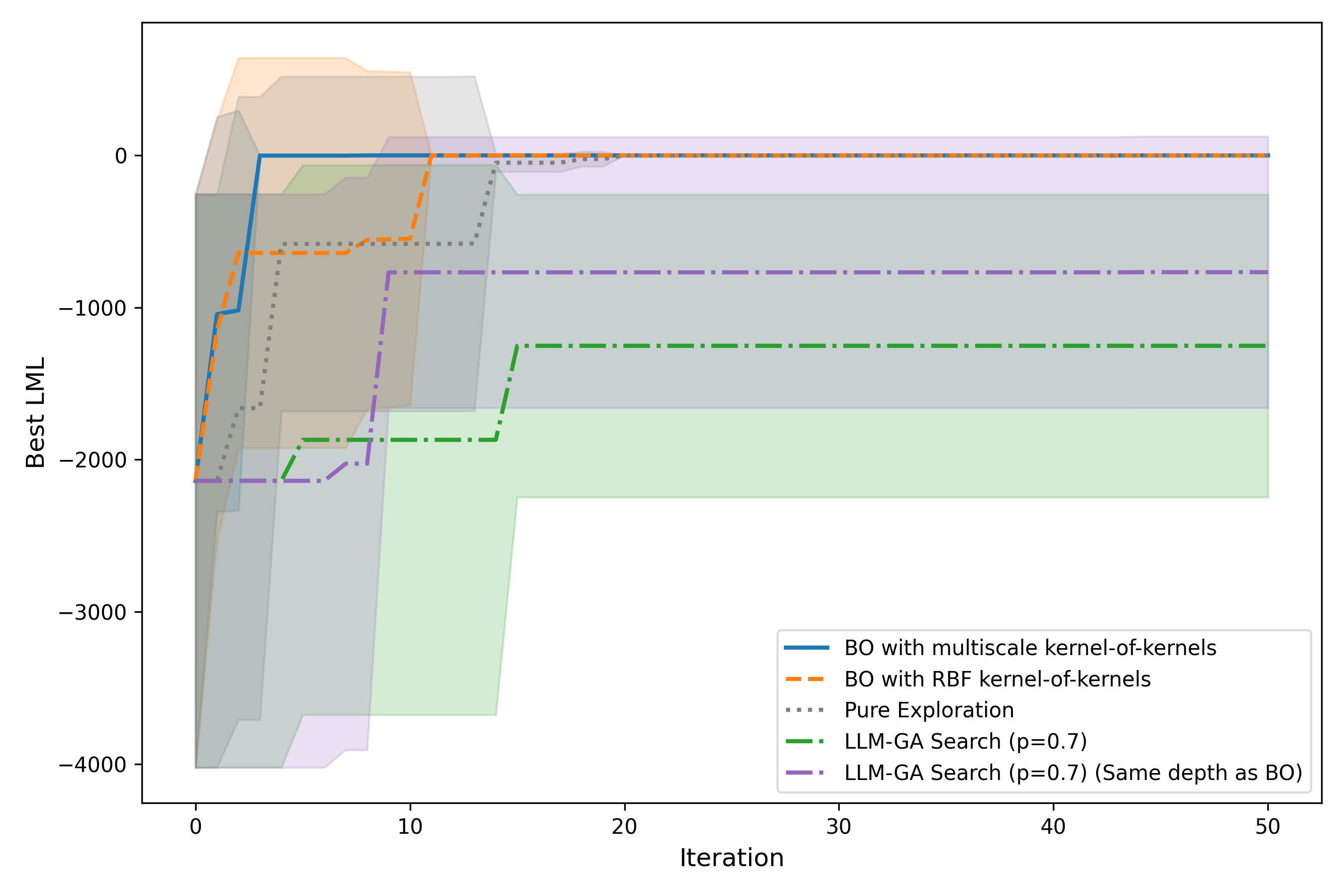} &
        \includegraphics[width=0.30\textwidth]{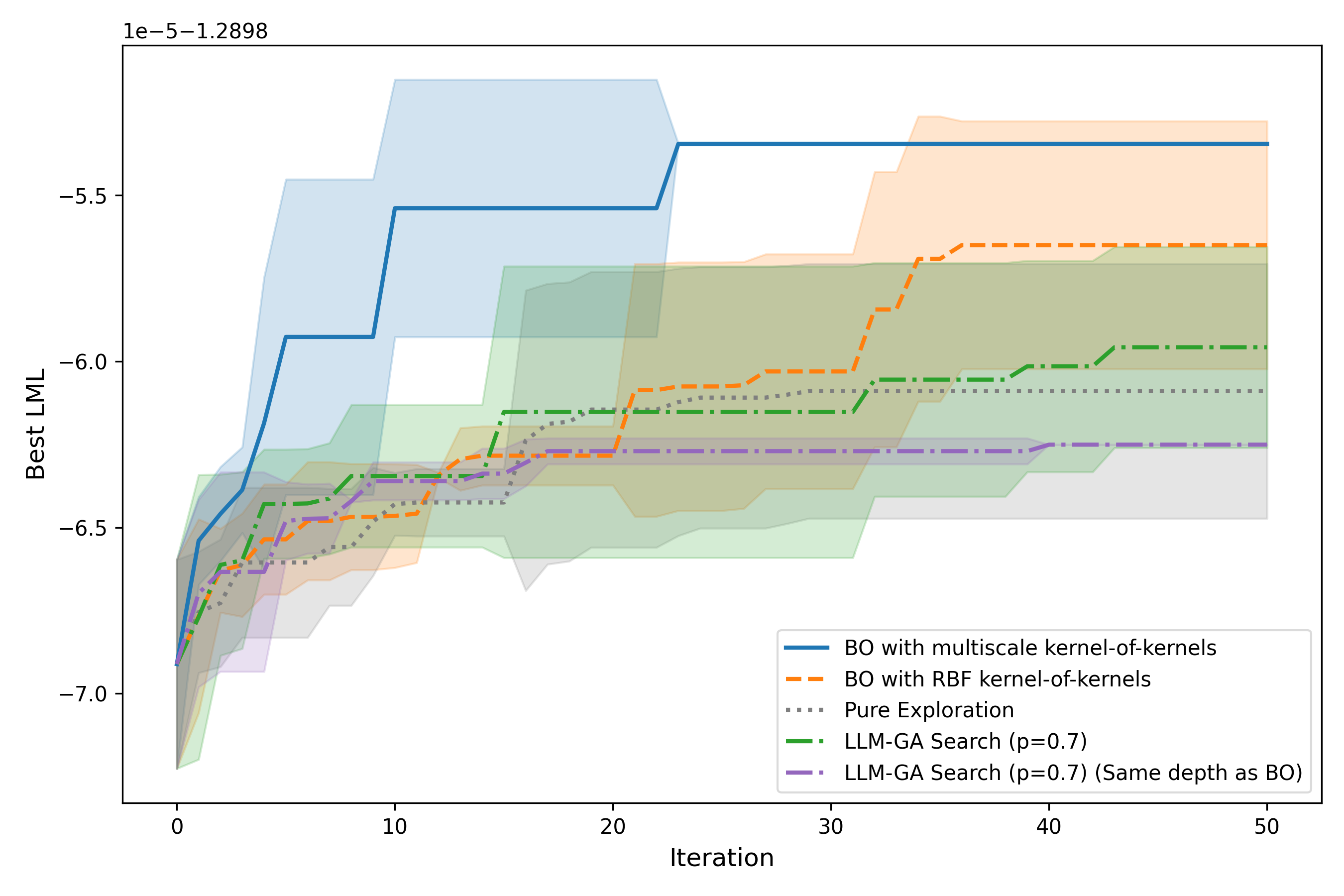} \\

        (a) Eggholder & (b) Ackley & (c) Dropwave \\[0.7em]

        \includegraphics[width=0.30\textwidth]{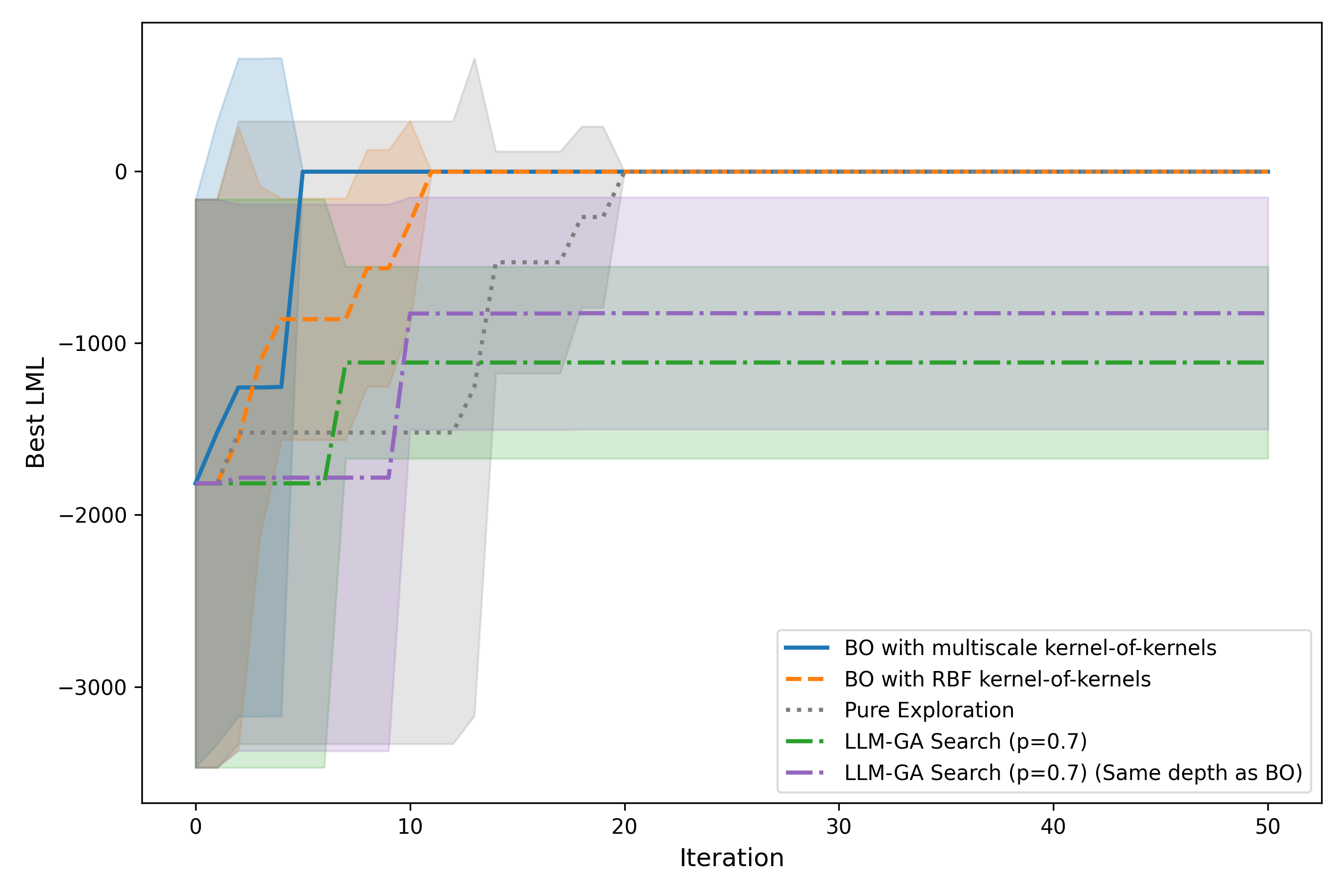} &
        \includegraphics[width=0.30\textwidth]{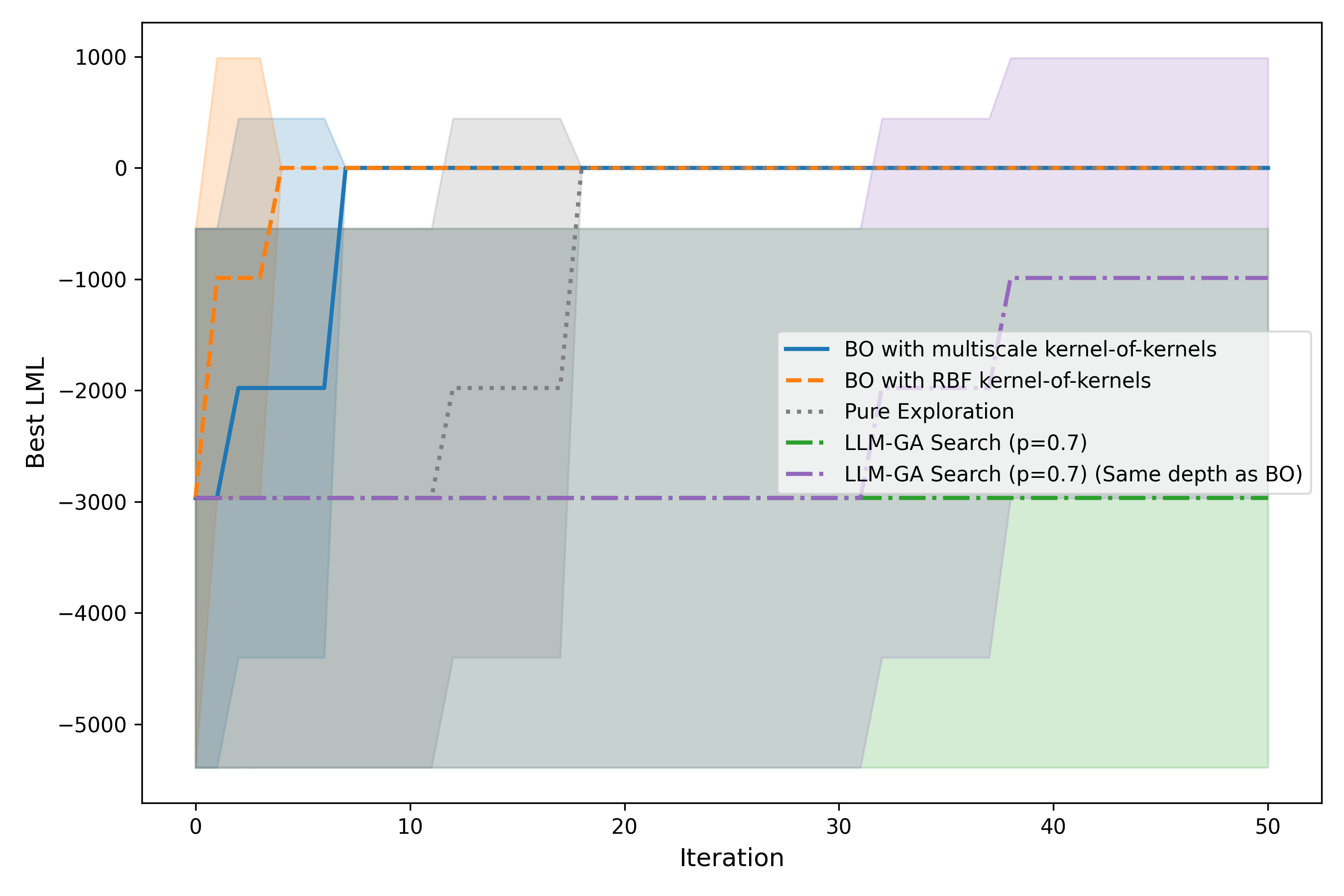} &
        \includegraphics[width=0.30\textwidth]{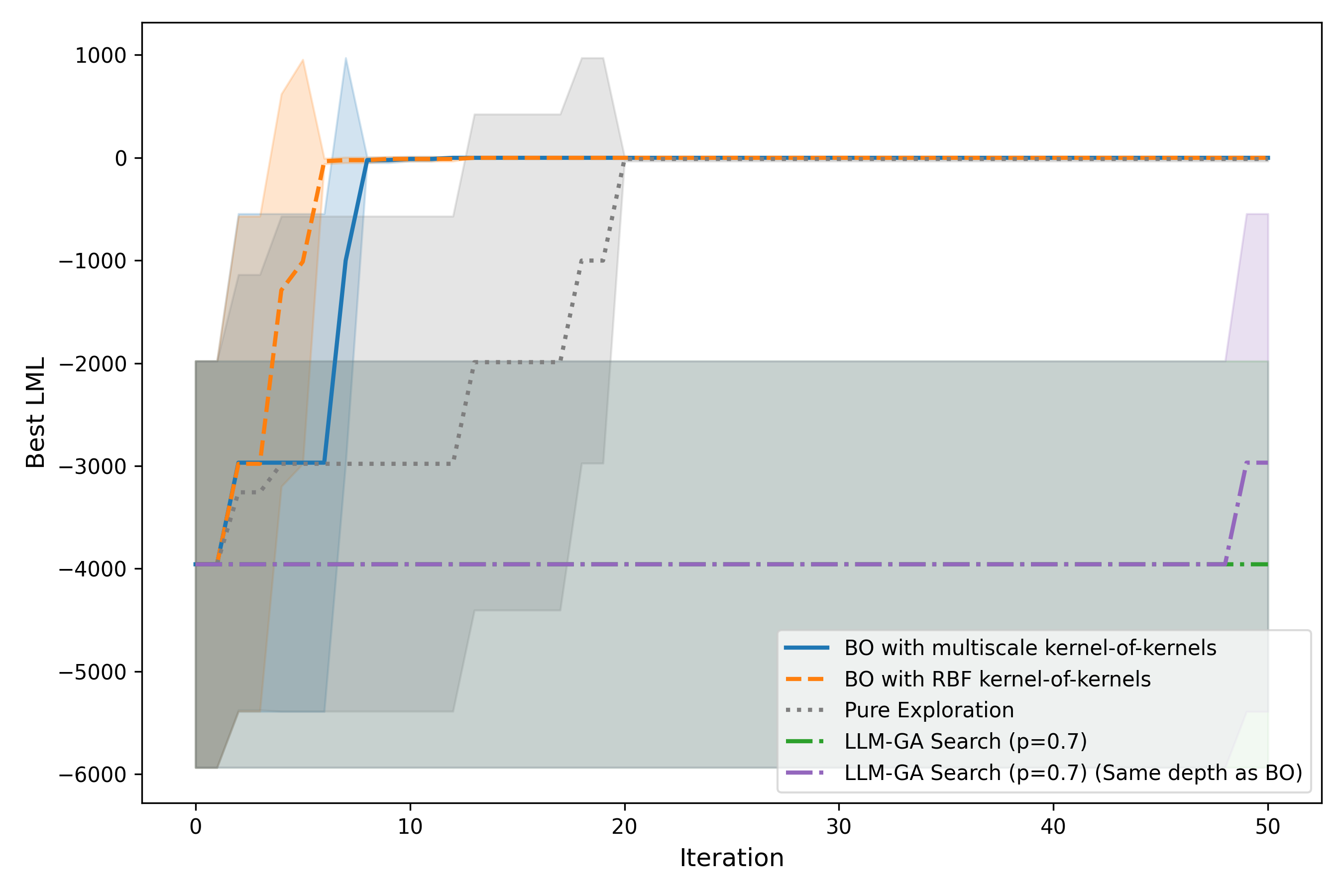} \\

        (d) Schwefel & (e) Rastrigin & (f) L\'evy \\[0.7em]

        \includegraphics[width=0.30\textwidth]{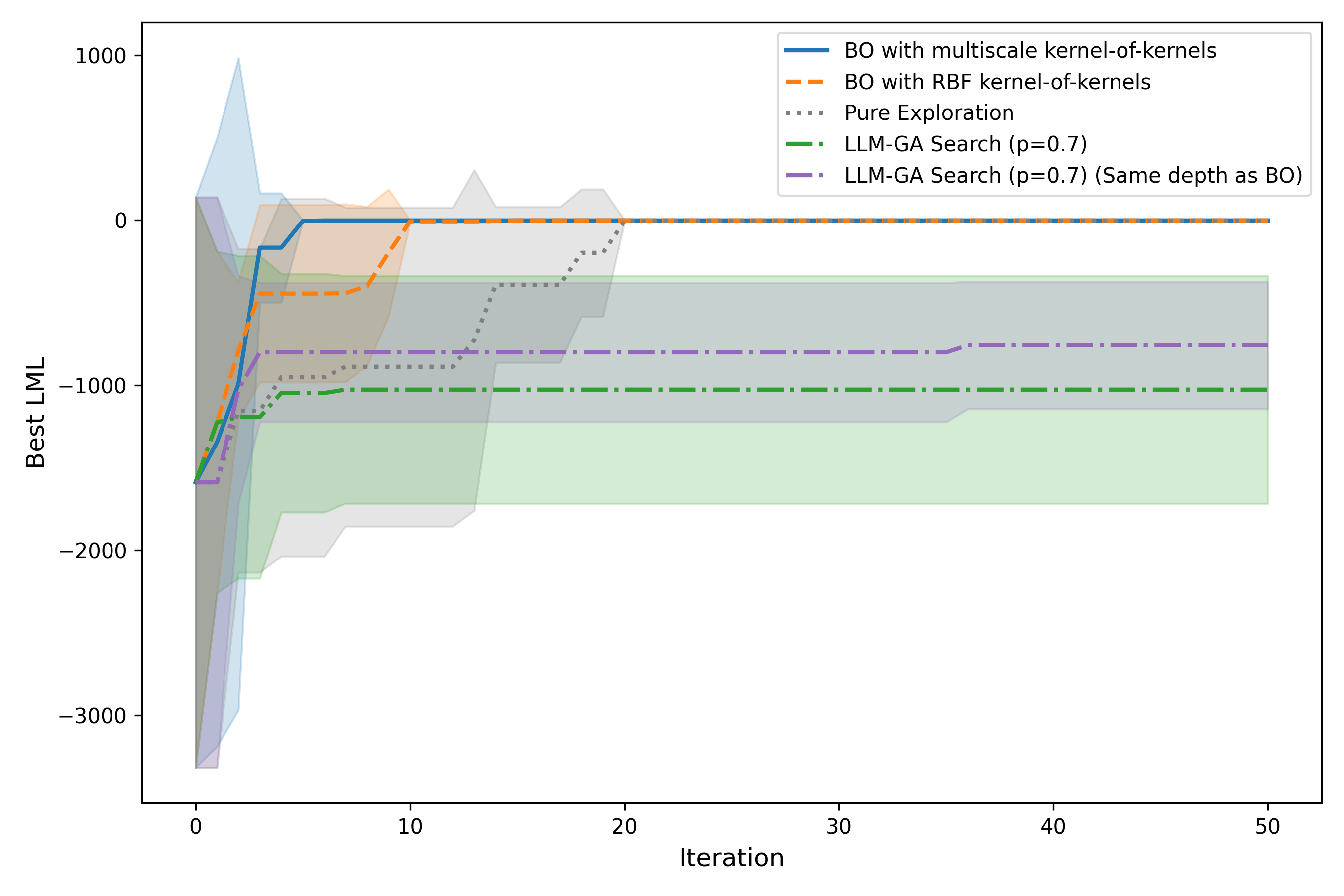} &
        \includegraphics[width=0.30\textwidth]{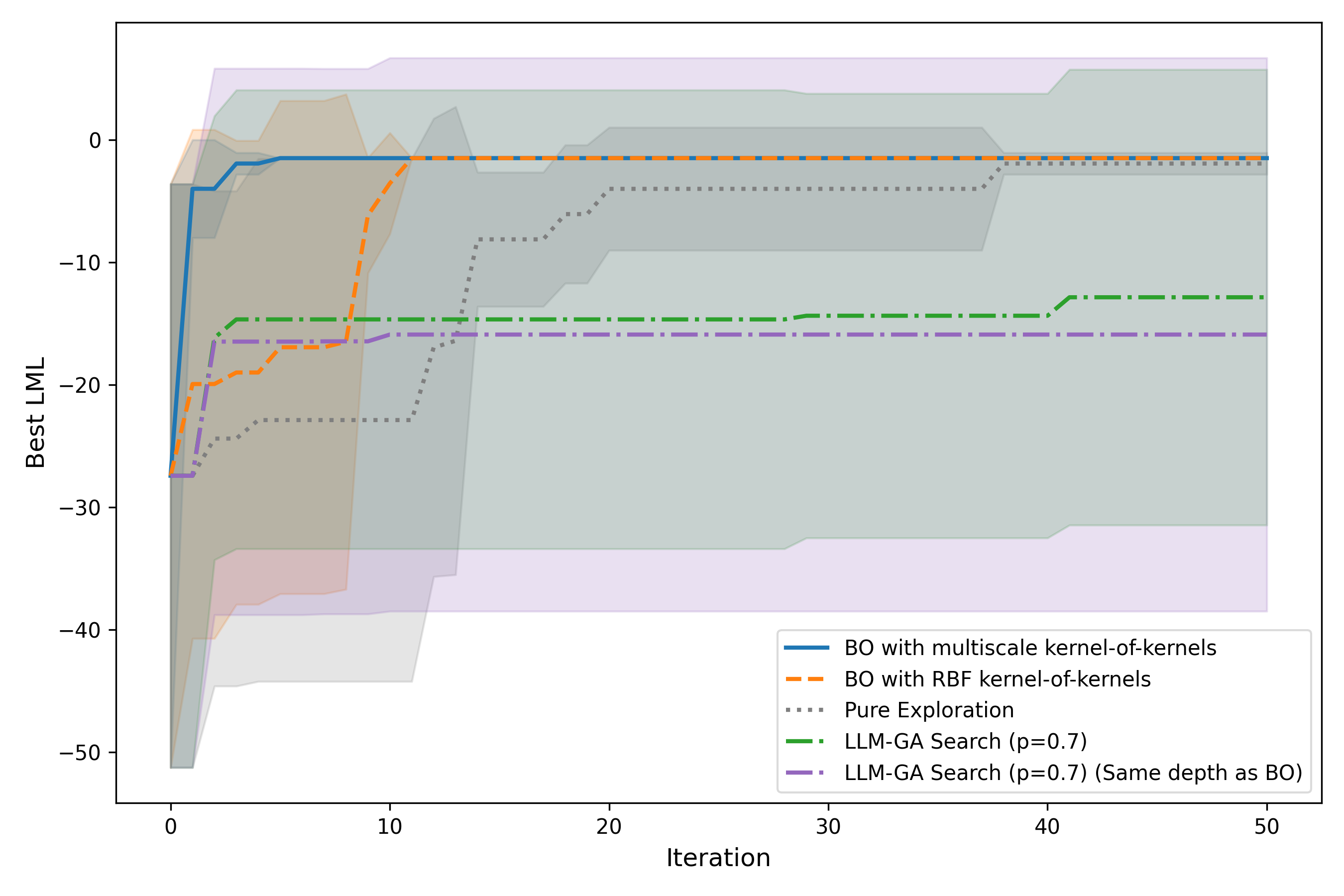} &
        \includegraphics[width=0.30\textwidth]{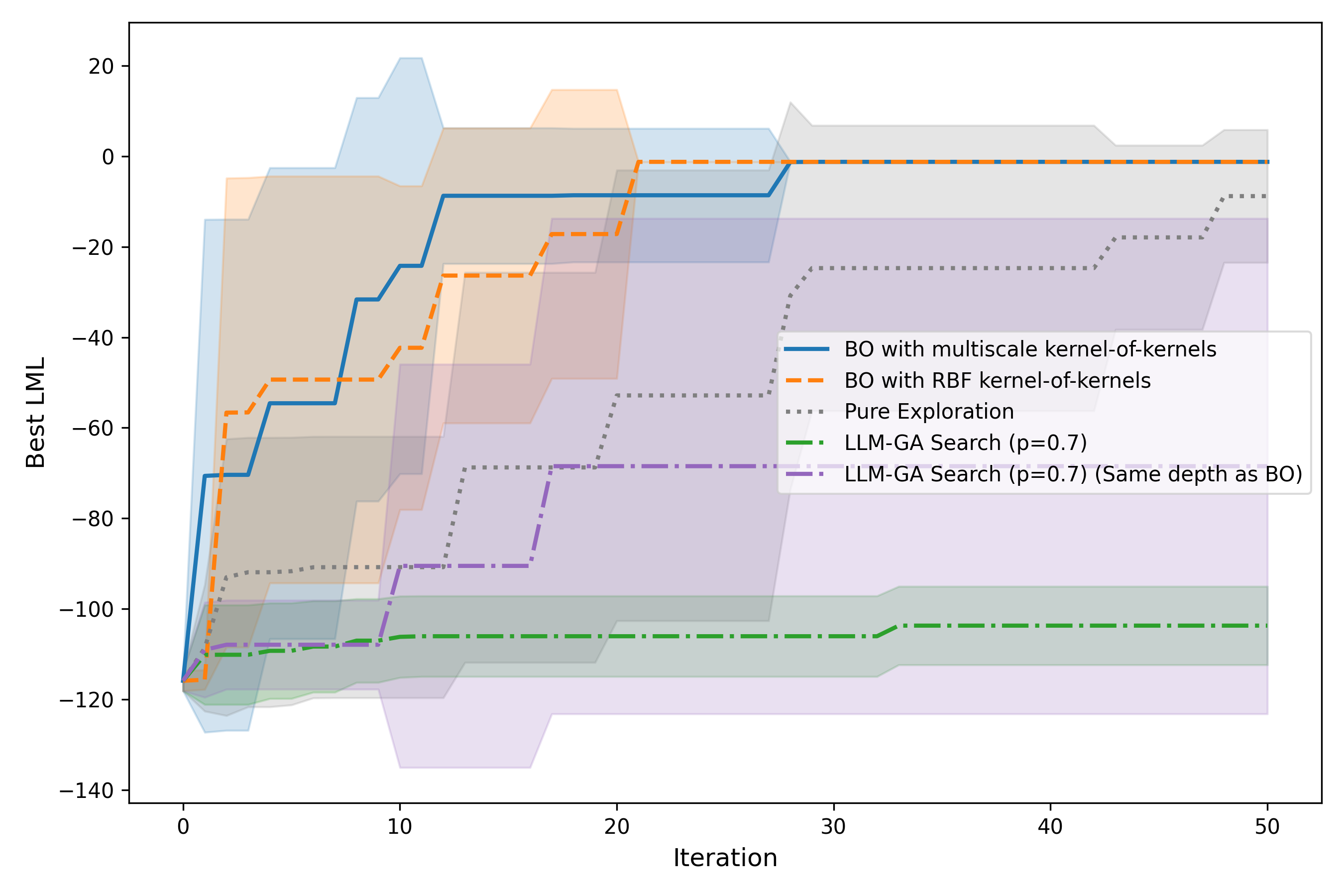} \\

        (g) Bukin & (h) Airline & (i) Mauna Loa CO$_2$ \\[0.7em]

        \multicolumn{3}{c}{
            \includegraphics[width=0.30\textwidth]{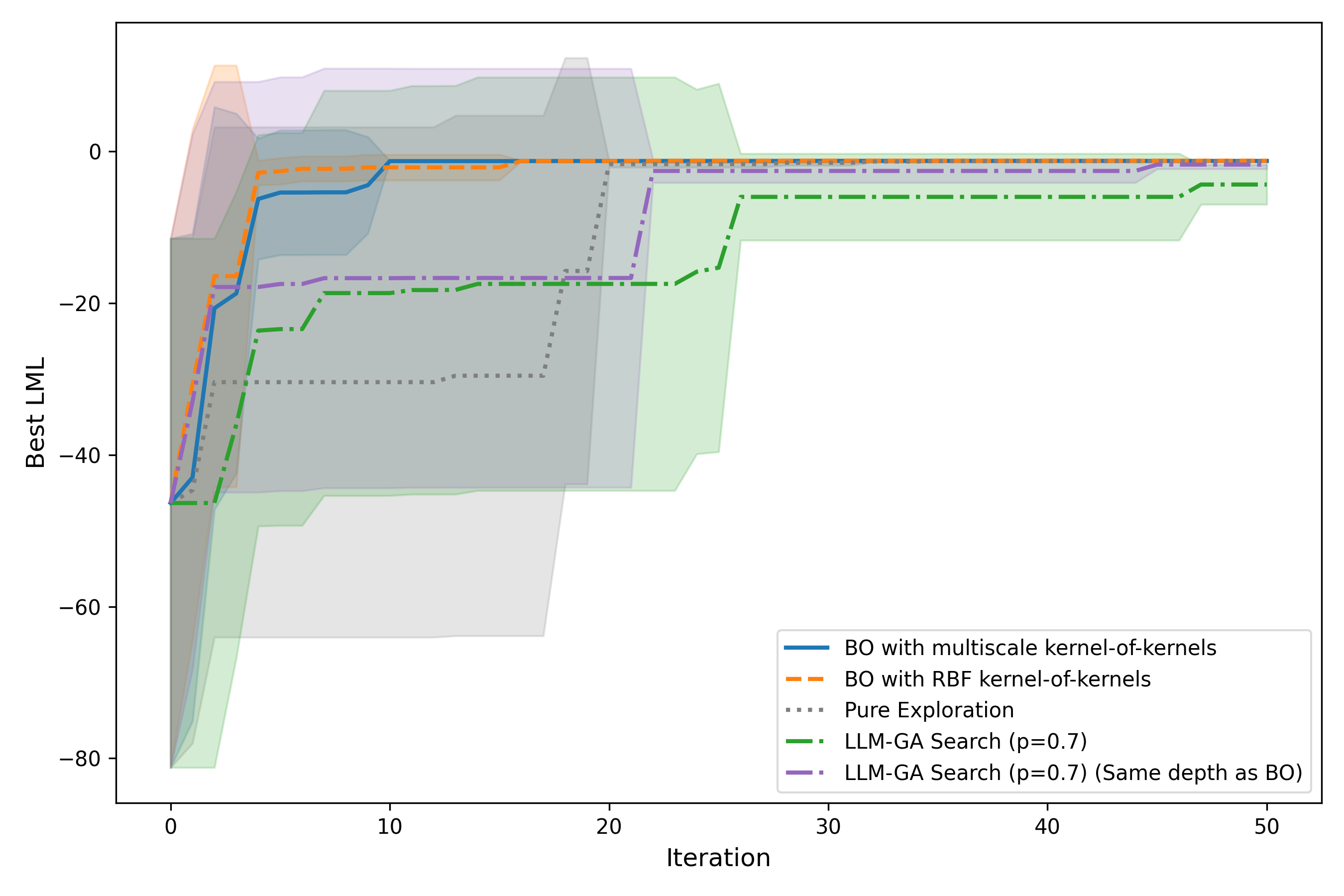}
        } \\
        \multicolumn{3}{c}{(j) Thermal History}
    \end{tabular}

    \caption{Performance of Multiscale kernel-of-kernel based Bayesian Optimization, RBF kernel-of-kernel based Bayesian Optimization, and Pure Exploration, and LLM-based Search (mutation probability = 0.7) across ten benchmark functions. Each subplot shows the best log marginal likelihood. Shaded regions indicate $\pm 1$ standard deviation across repeated runs. The proposed Multiscale kernel-of-kernel based BO converges faster and reaches superior solutions across all tasks.}
    \label{fig:benchmark_panel}
\end{figure*}

We benchmark kernel selection via BO on synthetic and real-world datasets using a shared MDS-embedded kernel space. The results compare multi-scale kernel-of-kernels BO, a single-RBF kernel-of-kernels baseline, random selection, and an LLM-GA search (detailed below), and conclude with case studies showing how optimized kernels improve prediction and downstream BO performance.

\subsection{Benchmarking Kernel Optimization with Synthetic Functions and Real-world Data}
To evaluate kernel selection via BO over the kernel-of-kernels space, we ran experiments on ten diverse functions: seven classical synthetic benchmarks (Eggholder \cite{whitley1996evaluating}, Ackley \cite{ackley2012connectionist}, Dropwave \cite{posypkin2017implementation}, Schwefel \cite{niewenhuis2023classical}, Rastrigin \cite{rastrigin1974systems}, L\'evy \cite{odili2018implementation}, and Bukin \cite{omae2024gaussian}) and three real-world time-series datasets (International Airline Passenger, Mauna Loa CO$_2$, and an in-house thermal history series). These problems span continuous, multimodal, deceptive, highly oscillatory, and nonstationary behaviors. We compare multi-scale kernel-of-kernels BO, single-RBF kernel-of-kernels BO, and random selection; LLM-GA results are reported in the next subsection. Each experiment was repeated multiple times to obtain statistically stable performance curves. The corresponding convergence trajectories, measured by best-observed log marginal likelihood, are summarized in Fig.~\ref{fig:benchmark_panel}.

On the synthetic benchmarks, BO over the kernel-selection space (embedded kernel manifold) shows consistently faster convergence and markedly reduced variance when using the multiscale kernel-of-kernels surrogate. For rugged and deceptive landscapes such as Eggholder and Schwefel, it rapidly escapes poor basins and approaches near-optimal regions within a few iterations, whereas the single-RBF kernel-of-kernels baseline progresses more slowly and often becomes trapped in local minima. Functions with oscillatory or periodic structure (Dropwave, Rastrigin) benefit from the kernel-manifold geometry: periodic and quasi-periodic kernels lie nearby in the embedding, enabling efficient selection of appropriate models. In contrast, the standard RBF model over-smooths these functions, yielding slower and less stable convergence.

Even for relatively smooth functions such as Lévy, the multiscale kernel-of-kernels BO shows improved convergence, reflecting its ability to select kernels that adapt to both global curvature and local irregularity. The Bukin function, with its sharp valleys and quasi-discontinuous structure, highlights the limitations of the single-RBF baseline, while the embedded geometry guides BO toward kernels capable of capturing nonstationary or asymmetric behavior. The aggregate effect across all synthetic tasks is visible in Fig.~\ref{fig:benchmark_panel}(a)--(g).

The three real-world time-series datasets pose different challenges, including long-range dependencies, seasonal components, irregular cycles, trend–seasonality interactions, and nonstationary structure. On the International Airline Passenger dataset, multiscale kernel-of-kernels BO identifies composite kernels that capture annual periodicity and long-term growth, yielding rapid improvements in log marginal likelihood. The Mauna Loa CO$_2$ dataset, which also exhibits strong annual periodicity with small sub-seasonal variations, is similarly well handled; kernel combinations such as periodic$\times$RBF or periodic+RQ naturally arise in the embedding space and are efficiently discovered through BO.

The in-house thermal history dataset is the most irregular and noisy of the real tasks, with abrupt thermal transitions and variable cycle durations. The multiscale kernel-of-kernels BO again outperforms the other methods, followed by the single-RBF kernel-of-kernels baseline using the same MDS embedded features, indicating that the embedding captures kernels capable of approximating nonstationary phenomena. While both baselines show limited improvement over random search for this dataset, the proposed method converges significantly faster and attains substantially higher marginal likelihood values. These results appear in Fig.~\ref{fig:benchmark_panel}(h)--(j), confirming the utility of the kernel-of-kernels representation on real, noisy, industrially relevant data.

Across all ten experiments, the multiscale kernel-of-kernels BO achieves faster convergence and lower variance than the single-RBF baseline and random selection, indicating that the learned kernel geometry provides actionable structure for search. These gains are consistent across synthetic and real-world tasks, supporting the generality and sample efficiency of the approach. Moreover, proximity in the geometric kernel space reflects functional similarity between GP priors, which is more informative than proximity in a purely symbolic space; this alignment allows BO to select kernels matched to each task’s structure.

\vspace{1em}
We next compare against an LLM-guided genetic algorithm to assess a symbolic-search alternative.

\subsection{Comparison with LLM-guided Genetic Algorithm (LLM-GA) search method}
We compared our BO-based kernel-structure optimization strategy with an LLM-guided genetic-algorithm (LLM-GA) search method and a random-search baseline. The LLM-GA is a modified CAKE framework~\cite{suwandi2025adaptive}, which has been reported to outperform prior kernel-search baselines by using an LLM to drive crossover and mutation and scoring candidates with log marginal likelihood (LML). In our implementation, we condition the LLM with two system prompts: one without depth restrictions and one with a depth constraint of $\leq 3$. The depth-constrained prompt limits search complexity and aligns with the grammar depth used to generate our kernel library. The prompts for crossover and mutation are listed below.

    \paragraph{\textbf{System Prompt (Unrestricted)}}"You are an expert in Gaussian processes and kernel design.\\
    Available base kernels: SE (Squared Exponential/RBF), PER (Periodic), RQ (Rational Quadratic)\\
    Available operators: + (addition), * (multiplication)\\
    Your task is to propose kernel expressions that maximize the log marginal likelihood (LML) on the observed data.\\
    Higher LML values indicate better fit to the data.\\
    IMPORTANT: Output format must be:\\
    Kernel: <kernel\_expression>\\
    Analysis: <your reasoning>\\
    Kernel expressions must use parentheses for compound operations, e.g., (SE + PER), (SE * RQ), ((SE + PER) * RQ)"
    
    \paragraph{\textbf{System Prompt (Depth Restricted)}}"You are an expert in Gaussian processes and kernel design. \\
    Available base kernels: SE (Squared Exponential/RBF), PER (Periodic), RQ (Rational Quadratic)\\
    Available operators: + (addition), * (multiplication)\\
    Your task is to propose kernel expressions that maximize the log marginal likelihood (LML) on the observed data.\\
    Higher LML values indicate better fit to the data.\\
    CRITICAL CONSTRAINT: The kernel expression depth must not exceed {max depth}.\\
    - Depth 1: Single base kernel (e.g., SE, PER, RQ)\\
    - Depth 2: One operation (e.g., (SE + PER), (SE * RQ))\\
    - Depth 3: Two operations (e.g., ((SE + PER) * RQ), (SE + (PER * RQ)))\\
    IMPORTANT: Output format must be:\\
    Kernel: <kernel\_expression>\\
    Analysis: <your reasoning>\\
    Kernel expressions must use parentheses for compound operations. Keep expressions simple and within the depth limit!"
    
    \paragraph{\textbf{Crossover Prompt}}"You are given two parent kernels and their LML fitness scores:\\
    Parent 1: \{parent1\} (LML: \{fitness1:.3f\})\\
    Parent 2: \{parent2\} (LML: \{fitness2:.3f\})\\
    Please propose a new kernel that combines the strengths of both parents and may achieve higher LML.\\
    You can use operators: \{operators\}\\
    \{depth constraint\}\\
    Output format:\\
    Kernel: <your\_kernel>\\
    Analysis: <brief explanation>"
    
    \paragraph{\textbf{Mutation Prompt}}"You are given a kernel and its LML fitness score:\\
    Current: \{kernel\} (LML: \{fitness:.3f\})\\
    Please propose a modified kernel that may achieve higher LML.\\
    You can replace base kernels with: \{base kernels\}\\
    Or add/modify operators: \{operators\}\\
    \{depth constraint\}\\
    Output format:\\
    Kernel: <your\_kernel>\\
    Analysis: <brief explanation>"\\

    \paragraph{\textbf{Depth Constraint}} If depth restriction of $\leq 3$, then the following prompt is given: "IMPORTANT: Keep kernel depth at most \{max\_depth\}. Avoid deeply nested expressions!" Else, empty prompt string is given.\\

For initial comparison, we evaluated multi-scale and single-RBF kernel-of-kernels BO alongside LLM-GA (both unrestricted and depth-restricted) and a random-search baseline. We swept six mutation probabilities (0.05, 0.1, 0.3, 0.5, 0.7, 0.9) and fixed the remaining LLM-GA parameters: one crossover per iteration, population size 6, and LLM temperature 0.7. The LLM used OpenAI's \texttt{gpt-4o-mini}. This sweep identifies the overall best mutation probability, which is $p = 0.7$, used in subsequent comparisons. Results at iteration 12 are summarized in Table~\ref{tab:lml_max_summary}.

\begin{table*}[h!]
    \caption{Summary of the Maximum Log Marginal Likelihood (Max LML) score recorded over 5 independent runs at iteration 12 for our proposed BO method with multi-scale RBF kernel-of-kernels and RBF kernel-of-kernels, pure exploration, and LLM-GA search on different standard benchmark functions.}
    \label{tab:lml_max_summary}
    \centering
    \fontsize{8}{10}\selectfont
    \setlength{\tabcolsep}{3pt} % Adjust spacing as needed
    \resizebox{450pt}{!}{
    % Define 7 columns: 1 for Function Name + 6 for Data
    % No vertical lines (|) are used.
    \begin{tabular}{l cccccc}
        \toprule
        
        % ==================================================
        % SECTION 1: BASELINES
        % ==================================================
        \multirow{2}{*}{\textbf{Function}} & \multicolumn{4}{c}{\textbf{Proposed BO methods}} & \multicolumn{2}{c}{\multirow{2}{*}{\textbf{Pure Exploration}}} \\
        \cmidrule(lr){2-5}
         & \multicolumn{2}{c}{\textbf{RBF-BO}} & \multicolumn{2}{c}{\textbf{Multiscale-RBF-BO}} & \\
        \midrule
        
        Ackley    & \multicolumn{2}{c}{-1.5 $\pm$ 0.0}   & \multicolumn{2}{c}{-1.5 $\pm$ 0.0}    & \multicolumn{2}{c}{-582.3 $\pm$ 1099.0} \\
        Bukin     & \multicolumn{2}{c}{-1.5 $\pm$ 0.0}   & \multicolumn{2}{c}{-1.5 $\pm$ 0.0}  & \multicolumn{2}{c}{-734.9 $\pm$ 705.1} \\
        Dropwave  & \multicolumn{2}{c}{-1.3 $\pm$ 0.0}            & \multicolumn{2}{c}{-1.3 $\pm$ 0.0}    & \multicolumn{2}{c}{-1.3 $\pm$ 0.0} \\
        Eggholder & \multicolumn{2}{c}{-263.7 $\pm$ 524.5}   & \multicolumn{2}{c}{-1.5 $\pm$ 0.0} & \multicolumn{2}{c}{-735.0 $\pm$ 679.3} \\
        Levy      & \multicolumn{2}{c}{-990.5 $\pm$ 1977.9}   & \multicolumn{2}{c}{-1.5 $\pm$ 0.0} & \multicolumn{2}{c}{-3037.9 $\pm$ 2339.9} \\
        Rastrigin & \multicolumn{2}{c}{-1.35 $\pm$ 0.01} & \multicolumn{2}{c}{-1.36 $\pm$ 0.01}  & \multicolumn{2}{c}{-1979.4 $\pm$ 2422.5} \\
        Schwefel  & \multicolumn{2}{c}{-1.5 $\pm$ 0.0}   & \multicolumn{2}{c}{-1.5 $\pm$ 0.0} & \multicolumn{2}{c}{-1520.4 $\pm$ 1812.5} \\
        
        % ==================================================
        % SECTION 2: LLM-GA (Unrestricted)
        % ==================================================
        \midrule
        & \multicolumn{6}{c}{\textbf{LLM-GA}} \\
        \cmidrule(lr){2-7}
        \textbf{Function} & \multicolumn{6}{c}{\textbf{Mutation Probability ($p$)}} \\
        \cmidrule(lr){2-7}
        & \textbf{0.05} & \textbf{0.1} & \textbf{0.3} & \textbf{0.5} & \textbf{0.7} & \textbf{0.9} \\
        \midrule
        
        Ackley    & -2139.1 $\pm$ 1883.3 & -2139.1 $\pm$ 1883.3 & -2139.1 $\pm$ 1883.3 & -2139.1 $\pm$ 1883.3 & -2139.1 $\pm$ 1883.3 & -2139.1 $\pm$ 1883.3 \\
        Bukin     & -998.5 $\pm$ 683.6 & -998.5 $\pm$ 683.6 & -998.5 $\pm$ 683.6 & -998.5 $\pm$ 683.6 & -998.2 $\pm$ 683.6 & -998.5 $\pm$ 683.6 \\
        Dropwave  & -1.3 $\pm$ 0.0 & -1.3 $\pm$ 0.0 & -1.3 $\pm$ 0.0 & -1.3 $\pm$ 0.0 & -1.3 $\pm$ 0.0 & -1.3 $\pm$ 0.0 \\
        Eggholder & -1752.4 $\pm$ 966.2 & -1682.1 $\pm$ 917.3 & -1708.2 $\pm$ 855.8 & -1794.5 $\pm$ 914.8 & -1744.2 $\pm$ 959.0 & -1533.8 $\pm$ 876.9 \\
        Levy      & -3957.4 $\pm$ 1977.9 & -3957.4 $\pm$ 1977.9 & -3957.4 $\pm$ 1977.9 & -3957.4 $\pm$ 1977.9 & -3957.4 $\pm$ 1977.9 & -3957.4 $\pm$ 1977.9 \\
        Rastrigin & -2968.4 $\pm$ 2422.5 & -2628.6 $\pm$ 2233.0 & -2968.4 $\pm$ 2422.5 & -2968.4 $\pm$ 2422.5 & -2968.4 $\pm$ 2422.5 & -2968.4 $\pm$ 2422.5 \\
        Schwefel  & -1816.3 $\pm$ 1654.2 & -1816.3 $\pm$ 1654.2 & -1782.9 $\pm$ 1591.1 & -1782.9 $\pm$ 1591.1 & -1112.7 $\pm$ 558.8 & -1782.9 $\pm$ 1591.1 \\

        % ==================================================
        % SECTION 3: LLM-GA (Depth <= 3)
        % ==================================================
        \midrule
        & \multicolumn{6}{c}{\textbf{LLM-GA (Search Space Identical to BO)}} \\
        \cmidrule(lr){2-7}
        \textbf{Function} & \multicolumn{6}{c}{\textbf{Mutation Probability ($p$)}} \\
        \cmidrule(lr){2-7}
        & \textbf{0.05} & \textbf{0.1} & \textbf{0.3} & \textbf{0.5} & \textbf{0.7} & \textbf{0.9} \\
        \midrule
        
        Ackley    & -2026.5 $\pm$ 1879.9 & -2053.9 $\pm$ 1842.4 & -1150.2 $\pm$ 1380.9 & -2139.1 $\pm$ 1883.3 & -881.0 $\pm$ 1044.3 & -1499.8 $\pm$ 1954.4 \\
        Bukin     & -984.6 $\pm$ 682.5 & -579.3 $\pm$ 480.3 & -965.7 $\pm$ 688.1 & -758.8 $\pm$ 386.1 & -565.6 $\pm$ 466.6 & -739.9 $\pm$ 385.1 \\
        Dropwave  & -1.3 $\pm$ 0.0 & -1.3 $\pm$ 0.0 & -1.3 $\pm$ 0.0 & -1.3 $\pm$ 0.0 & -1.3 $\pm$ 0.0 & -1.3 $\pm$ 0.0 \\
        Eggholder & -1961.5 $\pm$ 1027.3 & -1979.3 $\pm$ 1006.2 & -2047.8 $\pm$ 1057.0 & -2023.0 $\pm$ 1052.1 & -1932.8 $\pm$ 1020.6 & -2039.6 $\pm$ 1052.8 \\
        Levy      & -2968.4 $\pm$ 2422.5 & -2968.4 $\pm$ 2422.5 & -3957.4 $\pm$ 1977.9 & -3957.4 $\pm$ 1977.9 & -3957.4 $\pm$ 1977.9 & -3957.4 $\pm$ 1977.9 \\
        Rastrigin & -2628.6 $\pm$ 2233.0 & -2968.4 $\pm$ 2422.5 & -2968.4 $\pm$ 2422.5 & -1979.4 $\pm$ 2422.5 & -1979.4 $\pm$ 2422.5 & -1979.4 $\pm$ 2422.5 \\
        Schwefel  & -1552.4 $\pm$ 1810.1 & -1782.8 $\pm$ 1591.1 & -1816.3 $\pm$ 1654.2 & -1782.9 $\pm$ 1591.1 & -1816.3 $\pm$ 1654.2 & -1782.9 $\pm$ 1591.1 \\
        
        \bottomrule
    \end{tabular}}
    \begin{tablenotes}
        \item Scores are reported as Mean $\pm$ Standard Deviation.
    \end{tablenotes}
\end{table*}

Table~\ref{tab:lml_max_summary} and Fig.~\ref{fig:benchmark_panel} show the same pattern: the single-RBF and multi-scale kernel-of-kernels BO achieves the highest or tied-best LML across most functions, while both LLM-GA variants trail by a wide margin and exhibit large variance. Performance is highly sensitive to the LLM-GA mutation probability, indicating unstable search dynamics and the need for careful tuning. The depth-restricted prompt yields some improvements but does not close the gap, suggesting that added expressiveness alone is not the limiting factor. Notably, even pure exploration (random kernel selection) generally outperforms the LLM-based approach under these settings, and the higher variance in LLM-GA results indicates weaker reproducibility. However, from Table~\ref{tab:lml_max_summary}, the mutation probability of 0.7 was observed to produce overall best LML across most functions similar to the results of CAKE framework~\cite{suwandi2025adaptive}. This mutation probability ($p = 0.7$) for detailed kernel optimization study for all the ten functions (seven synthetic benchmarks and three real-world datasets) as shown in Fig.~\ref{fig:benchmark_panel}.

The consistent under-performance of LLM-GA relative to our geometric approach highlights a fundamental mismatch between symbolic exploration and the functional geometry of Gaussian Process priors. While CAKE framework~\cite{suwandi2025adaptive} used Bayesian Information Criteria (BIC) for kernel evolutionreported success using the Bayesian Information Criterion (BIC), this metric is a conservative point-estimate proxy that favors simplicity and is significantly less sensitive to the underlying covariance structure~\cite{kass1995bayes, minka2000bayesian}. Our results demonstrate that when the objective is shifted to the LML, the optimization landscape becomes far more rugged and sensitive to structural changes~\cite{lofti2022bayesian}. Because LLMs operate via autoregressive token prediction rather than geometric reasoning, they are highly prone to "syntactic brittleness"~\cite{yuksel2025evolattice, williamson2025syntactic}. A single string-level mutation, such as swapping a product for a sum, can cause a catastrophic jump in the model's behavior. Rather than iteratively refining successful functional building blocks (genetic schemas), the stochastic nature of the search often leads to their catastrophic disruption~\cite{whitley1994genetic}. This instability is further exacerbated by the GA’s sensitivity to hyperparameters like the mutation probability ($p$), which leads to the high variance and search redundancy observed in Table~\ref{tab:lml_max_summary}. Ultimately, without the neighborhood-awareness provided by our MDS-embedded manifold, the LLM-GA struggles to preserve the very building blocks necessary for incremental progress in complex kernel space.

\begin{figure}[t]
  \centering
  \includegraphics[width=0.48\textwidth]{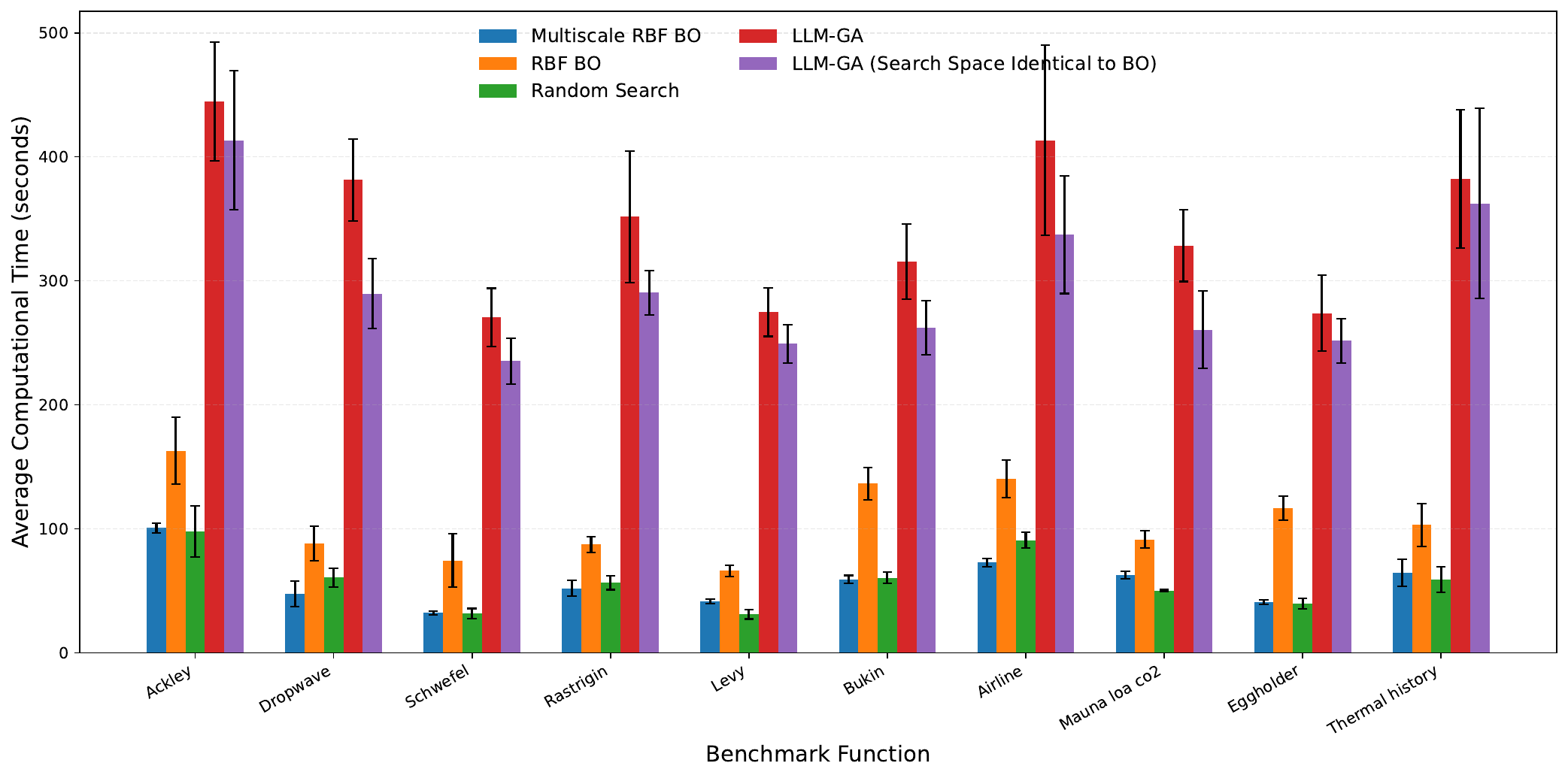}
  \caption{
  Comparison of average computational time (seconds) for Multiscale RBF BO, RBF BO, Random Search, and LLM-GA variants across ten benchmark functions over 50 iterations.
  Error bars denote $\pm 1$ standard deviation over five independent runs.
  LLM-GA incurs substantially higher computational overhead, while BO-based methods remain consistently efficient.
  }
  \label{fig:computational_time}
\end{figure}

We also compared the computational efficiency of our proposed BO-based kernel-structure optimization strategies against the LLM-GA workflow and pure exploration, as shown in Figure~\ref{fig:computational_time}. The results indicate that the LLM-GA workflow incurs substantially higher computational time, requiring 3.4 to 5.7 times the execution time of our proposed BO methods and pure exploration on average. This overhead can be attributed to the inference latency inherent in querying the LLM for prompt-based kernel generation and the cumulative delay of repeated API calls. Having established these benchmarking trends, we next demonstrate performance on materials case studies.

\subsection{Budgetary impact of inline kernel optimization during BO}

While figure \ref{fig:computational_time} reports the absolute computational overhead of kernel optimization, its practical impact must be interpreted relative to the cost of objective evaluations in real Bayesian optimization workflows. In many scientific and engineering applications, a single BO iteration corresponds to an expensive experiment or simulation, often ranging from several minutes to hours or days.

For example, we can consider a realistic setting in which BO is initialized with 50 observations from a black-box objective. Performing kernel optimization on this static dataset incurs an upfront cost of approximately 1--2 minutes. Subsequent BO iterations may each require on the order of 10 minutes (or more) per evaluation. After 10 BO iterations, the dataset grows to 60 points, and re-optimizing the kernel again incurs a comparable marginal cost. In this regime, kernel optimization contributes roughly 2 minutes of overhead relative to approximately 100 minutes spent on objective evaluations, corresponding to a $\sim$2\% increase in total runtime.

\begin{figure}[t]
    \centering
    \includegraphics[width=0.9\columnwidth]{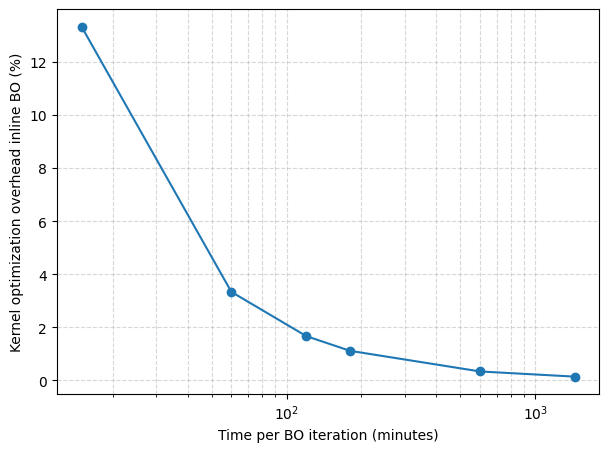}
    \caption{Illustrative scaling of the relative computational overhead introduced by inline kernel optimization as a function of the cost per Bayesian optimization (BO) iteration. The curve shows the fraction of total runtime attributable to kernel optimization under a simple budget model with periodic kernel re-optimization as the dataset grows. As the cost of individual BO evaluations increases—from minutes to hours or days—the relative overhead from kernel optimization decays rapidly and becomes negligible. This figure is schematic and intended to illustrate scaling behavior rather than report additional benchmark measurements.}
    \label{fig:kernel_overhead_scaling}
\end{figure}

As the cost of individual BO evaluations increases, e.g., to 30 minutes, several hours, or even days of experiments, the relative overhead introduced by kernel optimization decreases rapidly and becomes effectively negligible, as illustrated in Fig.~\ref{fig:kernel_overhead_scaling}. Consequently, inline kernel optimization does not meaningfully inflate the total BO budget and is well justified in settings where surrogate fidelity is critical to sample efficiency.

\subsection{Case Study 1: Melt-Pool Geometry Data with Process variability in AM process}

To demonstrate applicability to real materials problems, we applied the framework to a two-dimensional process-parameter space defined by laser power and scan speed in additive manufacturing.  
Melt-pool width, computed using the Thermo-Calc \textregistered~\cite{andersson2002thermo} Additive Manufacturing (TCAM) module, served as the output quantity of interest.

\begin{figure}[h]
  \centering
  \includegraphics[width=0.4\textwidth]{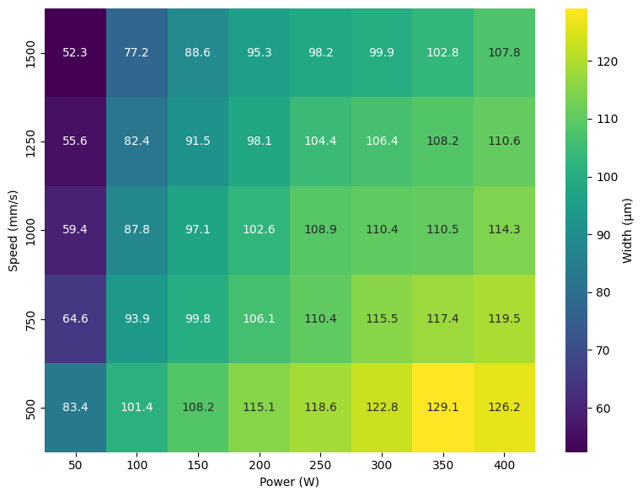}
  \caption{Experimental TCAM data showing melt-pool width as a function of laser power and scan speed.}
  \label{fig:tcam_data}
\end{figure}

\begin{figure}[ht]
  \centering
  \includegraphics[width=0.4\textwidth]{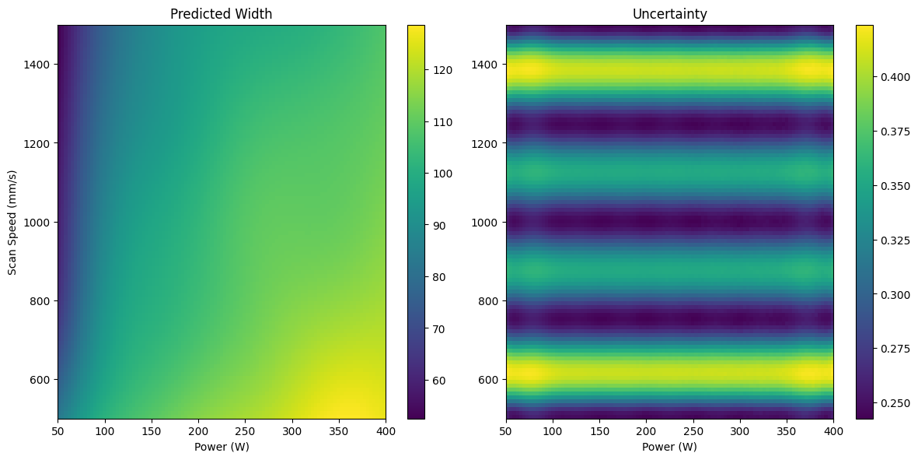}
  \caption{GP prediction using a default kernel from BoTorch. Left: predicted melt-pool width. Right: posterior uncertainty. The default kernel fails to capture complex interactions and exhibits large uncertainties where data are sparse.}
  \label{fig:default_kernel}
\end{figure}

Figure~\ref{fig:tcam_data} presents the measured melt-pool widths across a $5\times5$ grid of power and speed settings.  
Training a GP with the default RBF kernel on this sparse dataset led to significant shortcomings (Fig.~\ref{fig:default_kernel}): the predictive mean lacked local fidelity and the uncertainty map showed extensive high-variance regions in sparsely sampled areas. 
This is a fundamental limitation of single-scale stationary kernels: they fail to reconcile sharp gradients and plateaus simultaneously and require prohibitively dense sampling to reduce uncertainty.

\begin{figure}[ht]
  \centering
  \includegraphics[width=0.4\textwidth]{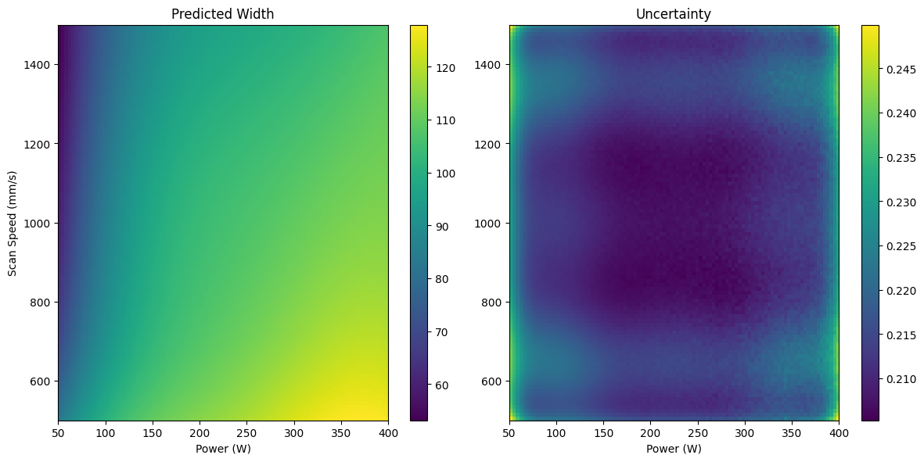}
  \caption{GP prediction using the optimized kernel. Left: predicted melt-pool width. Right: posterior uncertainty. The optimized kernel yields low, nearly uniform uncertainty and accurately reproduces the TCAM data.}
  \label{fig:optimized_kernel}
\end{figure}

The optimized kernel substantially alleviated these issues (Fig.~\ref{fig:optimized_kernel}).  
Despite the same sparse training set, the predicted width field closely matched the TCAM measurements.  
The predictive uncertainty was nearly uniform and low across the input domain.  
This suggests that the optimized composite kernel can flexibly represent the underlying physical process, capturing both smooth trends and subtle interactions between power and speed without overfitting.

From an engineering perspective, these results are important.  
Manufacturing process design often demands confidence in regions of the parameter space that have not been experimentally explored.  
High uncertainty in those regions can translate into costly trial-and-error.  
By embedding the entire symbolic kernel library in a Hellinger-informed MDS space, the framework automatically selected a kernel whose inductive bias best matches the multi-scale physics of laser-matter interaction, providing reliable predictions even far from the training data.

This example shows the broader value of the kernel-of-kernels approach. By treating kernels as points in a probability-informed geometric manifold and optimizing over this continuous space, we automate the discovery of expressive kernels that substantially improve predictive accuracy and uncertainty quantification in real engineering applications. The result is more reliable extrapolation in sparsely sampled regions and reduced uncertainty without requiring denser sampling, because the selected kernels align with multi-scale physics rather than symbolic complexity.
\subsection{Case Study 2: Bayesian Optimization with the Optimized Kernel}
To assess how kernel optimization affects Bayesian optimization performance, we begin with single-objective benchmarks.

For each test function, we first used the kernel-of-kernels framework to identify its best composite kernel (e.g., RBF$\times$(RQ+RQ) for Ackley). We then treated the test function itself as the BO objective and compared two surrogates: the selected composite kernel versus a standard RBF kernel. The RBF kernel is a common default in BO because it is smooth, stationary, and broadly effective across many problems. Both methods used identical initialization, acquisition strategy, and evaluation budgets, differing only in the surrogate kernel. Performance is reported as best objective value vs. iteration, averaged over multiple randomized runs.

\begin{figure}[t]
    \centering
    \begin{subfigure}[t]{0.4\textwidth}
        \centering
        \includegraphics[width=\linewidth]{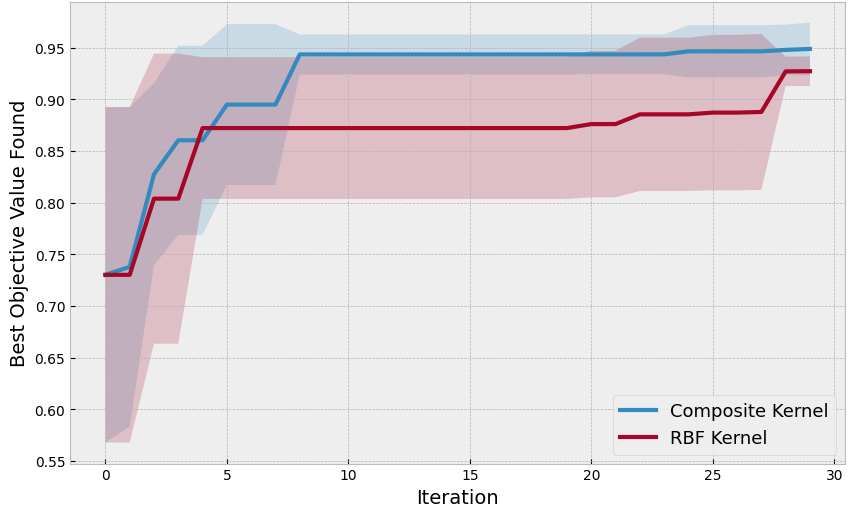}
        \caption{Single-objective benchmark with dropwave function}
        \label{fig:dropwave-bo}
    \end{subfigure}
    \hfill
    \begin{subfigure}[t]{0.4\textwidth}
        \centering
        \includegraphics[width=\linewidth]{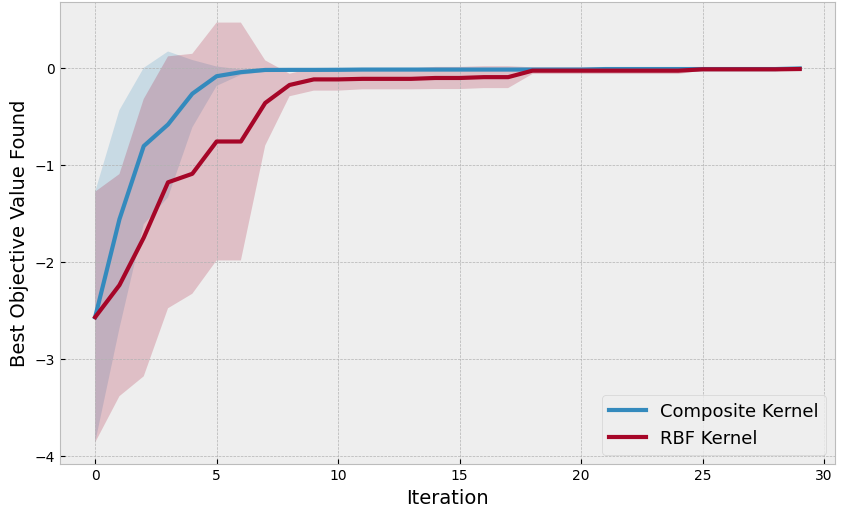}
        \caption{Single-objective benchmark with ackley function}
        \label{fig:ackley-bo}
    \end{subfigure}

    \caption{Comparison of Bayesian optimization performance using optimized composite kernels and a standard RBF kernel on single-objective benchmark functions. Results are averaged over multiple randomized runs. Solid lines indicate the mean best objective value found, while shaded regions denote $\pm 1$ standard deviation. The optimized kernel consistently achieves faster convergence and improved robustness across benchmark landscapes.}
    \label{fig:bo_single_objective}
\end{figure}

As shown in Fig.~\ref{fig:bo_single_objective}, BO with the task-specific composite kernel consistently outperforms the RBF baseline across both benchmarks. For Dropwave (Fig.~\ref{fig:dropwave-bo}), it reaches near-best observed values within the first few iterations, whereas the RBF kernel converges more slowly and with higher variability. A similar trend appears for Ackley (Fig.~\ref{fig:ackley-bo}), where the composite kernel escapes poor initial regions faster and reaches stronger objective values with reduced uncertainty.

These results indicate that the benefits of kernel optimization extend beyond surrogate quality metrics and directly translate into more efficient objective-space exploration. By selecting kernels whose inductive biases better align with the underlying landscape geometry, the proposed framework allows Bayesian optimization to balance exploration and exploitation more effectively than fixed-kernel baselines. Importantly, these improvements are achieved without modifying the acquisition strategy or optimization budget, confirming that the observed gains arise solely from the geometry-aware kernel selection enabled by the kernel-of-kernels approach.

In addition to benchmark studies, we performed Bayesian optimization directly on
physically meaningful printability objectives derived from Thermo-Calc's\textregistered~\cite{andersson2002thermo} 
Additive Manufacturing (TCAM) simulations. Here, \emph{printability} refers to
defect-free processing windows defined by melt-pool geometry criteria that avoid
lack of fusion, balling, and keyholing~\cite{johnson2019assessing,sheikh2024automated,sheikh2025high}.
We compared Bayesian optimization driven by task-specific optimized kernels,
obtained using the proposed kernel-of-kernels framework, against a standard
RBF-kernel baseline. Both approaches employed identical initialization strategies,
acquisition functions, and evaluation budgets, ensuring that any observed
performance differences arise solely from the surrogate kernel choice.

The objectives correspond to three defect criteria in laser powder bed fusion: lack of fusion, balling, and keyholing. Each is expressed as a melt-pool geometry inequality (e.g., $L - 3W < 0$). We treat these constraints as a multi-objective problem with soft penalties, where each violation is minimized. This enables continuous trade-off exploration, uncertainty-aware decision-making, and compatibility with hypervolume metrics.

\begin{figure}[t]
    \centering
    \includegraphics[width=0.4\textwidth]{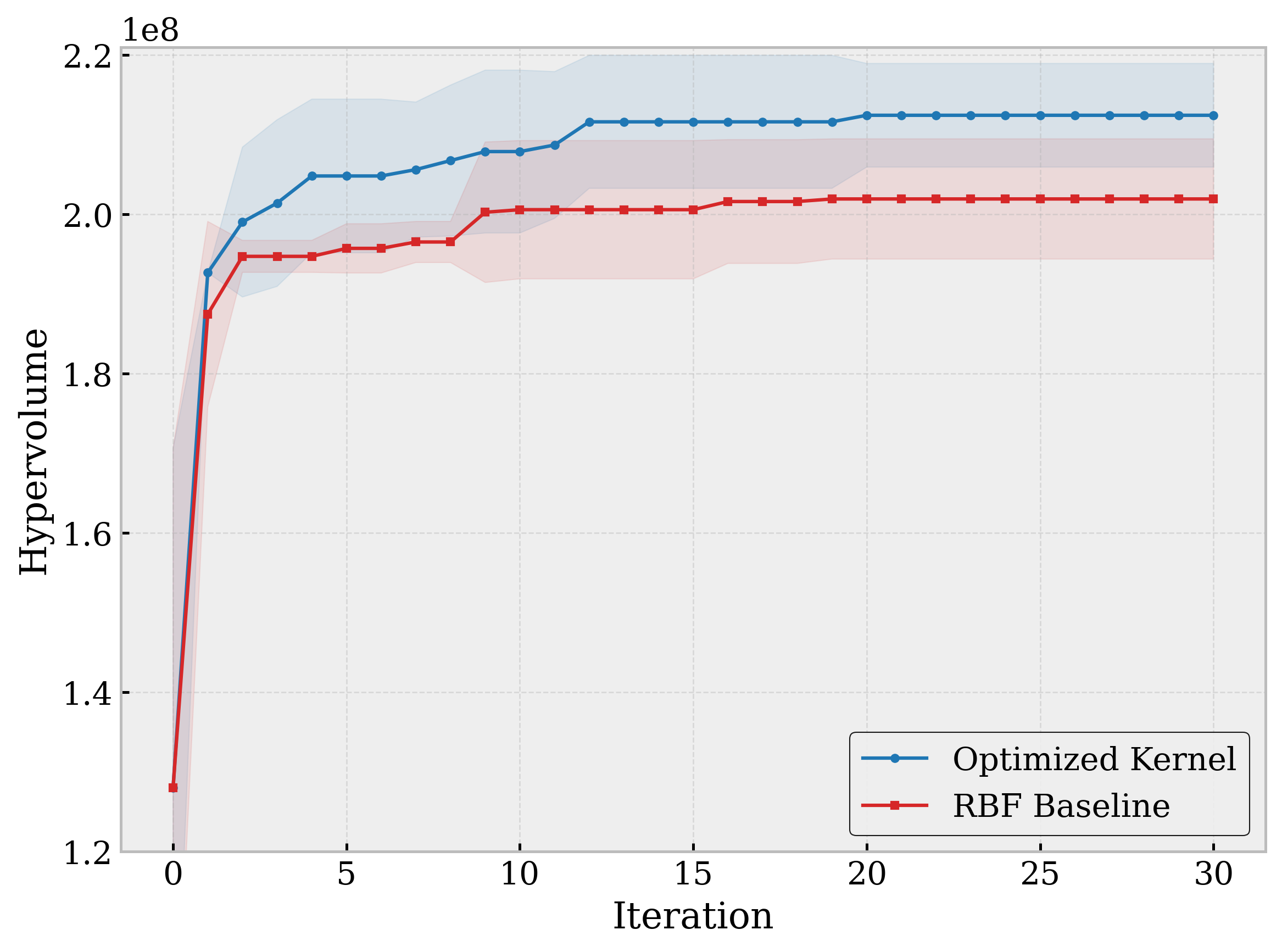}
    \caption{Evolution of the dominated hypervolume during multi-objective Bayesian
    optimization of TCAM-based printability criteria. Results are averaged over
    five independent runs. Solid lines indicate the mean hypervolume, while shaded
    regions denote $\pm 1$ standard deviation. Bayesian optimization using optimized
    composite kernels consistently achieves faster hypervolume growth and higher
    final hypervolume compared to the RBF baseline.}
    \label{fig:tcam_hypervolume}
\end{figure}

As shown in Fig.~\ref{fig:tcam_hypervolume}, BO with task-specific optimized kernels achieves faster hypervolume growth and higher final hypervolume than the RBF baseline. This indicates more effective simultaneous reduction of all three defect modes. The optimized-kernel approach also shows improved sample efficiency and reduced variability across runs, highlighting the robustness of the learned kernel representation.

These results confirm that the proposed kernel-optimization framework yields tangible
benefits at the decision level, enabling Bayesian optimization to more rapidly
identify processing conditions that satisfy multiple competing manufacturability
constraints. Although the problem is constraint-driven, treating defect criteria as soft objectives lets the optimizer explore infeasible regions early on and converge more reliably to high-quality feasible solutions than fixed-kernel baselines.

All experiments are reproducible using the publicly released
Kernels-BO codebase~\cite{islam_2026_18234227}.

\section{Conclusion}
This work establishes a geometric foundation for automated kernel discovery by optimizing kernels directly over their induced geometry. We construct divergence-based distances between GP priors, correct their curvature to recover Euclidean structure, and embed the resulting geometry with multidimensional scaling. This converts a combinatorial kernel library into a continuous manifold that can be explored efficiently by Bayesian Optimization. The approach provides a principled alternative to symbolic search or hyperparameter tuning, enabling BO to operate on representations that reflect functional behavior rather than algebraic form.

A central contribution is showing that, once regularized with monotone transformations, the kernel manifold admits a low-distortion Euclidean embedding where geometric proximity aligns with similarity of GP priors. This embedding lets acquisition functions reason about neighborhoods and uncertainty in kernel space, yielding a reproducible, data-adaptive mechanism for selecting kernels whose inductive biases match the task.

The optimization framework is general: it does not depend on a specific grammar, parameterization, or model class. It accommodates richer libraries, domain-specific priors, or hierarchical kernel constructions without changing the pipeline. By separating kernel geometry from syntax, it provides a foundation for automated nonstationary modeling, deep kernel learning, multi-fidelity models, and structured scientific simulators.

Beyond kernel discovery, this geometric perspective offers a general recipe for navigating discrete, structured model spaces. When candidates are symbolic objects, e.g., graphs, grammars, programs, or workflows, the key is to define distances that reflect functional similarity rather than syntactic overlap. In materials design, process routes or microstructure pathways can be compared through simulated responses, uncertainty profiles, or performance distributions. Once such task-aware distances are defined, the geometry provides a meaningful notion of neighborhoods, trade-offs, and uncertainty, allowing continuous optimization tools to guide search in otherwise combinatorial spaces. The kernel manifold is one concrete instance of this idea, but the same principle can support discovery in other structured domains where behavior, not form, is the relevant notion of similarity.

Overall, this work shows that kernel geometry provides a stable substrate for kernel discovery and opens a path toward scalable, interpretable model search. By tying proximity to functional behavior, it reduces model-design burden and points to broader uses in structured inference and scientific modeling.

\section*{Code and Data Availability}

The code used to generate the results in this study is openly available at
\url{https://github.com/shafiqmme/Kernels-BO}.
A permanent, versioned archive of the code is available via Zenodo
at \url{https://doi.org/10.5281/zenodo.18234227}.

All benchmark datasets used in this work are publicly available and are cited in the manuscript.
The additive manufacturing case-study data were generated using Thermo-Calc\textregistered{}
and are available from the corresponding author upon reasonable request, subject to licensing restrictions.

% ----------------------------------------------------
\section*{Acknowledgements}
The authors would like to acknowledge NSF for the financial support to grant No. NSF-2323611 (DMREF: Optimizing Problem formulation for prinTable refractory alloys via Integrated MAterials and processing co-design (OPTIMA)). RA also acknowledges the support from  the Army Research Laboratory through Cooperative Agreement Number W911NF-22-2-0106 as part of the HTMDEC program (BIRDSHOT Center). Computations were carried out at the Texas A\&M High Performance Research Computing (HPRC) facility.

% APPENDIX
% -----------------------------------------------------
%\appendix

%\section{Appendix}
%Additional proofs, derivations, and supplementary figures will be included here.

% -----------------------------------------------------
% BIBLIOGRAPHY
% -----------------------------------------------------
\bibliography{ref}
\bibliographystyle{ieeetr}

\end{document}